\documentclass[smallextended]{svjour3}       % onecolumn (second format)
\smartqed  % flush right qed marks, e.g. at end of proof
\usepackage{times}
\usepackage{epsfig}
\usepackage{graphicx}
\usepackage{amsmath}
\usepackage{amssymb}
\usepackage{subfigure}
\usepackage[abs]{overpic}
\usepackage{color}
\usepackage{rotating}

\usepackage[pagebackref=true,breaklinks=true,letterpaper=true,colorlinks,bookmarks=false]{hyperref}
\begin{document}

\definecolor{darkgray}{gray}{0.0}

\newcommand{\Ra}[1]{{\color{darkgray}#1}}
\newcommand{\RB}[1]{{\color{black}#1}}

% Include other packages here, before hyperref.

% If you comment hyperref and then uncomment it, you should delete
% egpaper.aux before re-running latex.  (Or just hit 'q' on the first latex
% run, let it finish, and you should be clear).

% Pages are numbered in submission mode, and unnumbered in camera-ready
%\ifcvprfinal\pagestyle{empty}\fi
%\begin{document}

%\baselinestretch

%%%%%%%%% TITLE
%\title{An Approximate Shading Model for Object Relighting}
\title{An Approximate Shading Model with Detail Decomposition for Object Relighting}

\author{Zicheng Liao
%{\tt\small zliao@zju.edu.cn}
\and
Kevin Karsch \and
Hongyi Zhang \and
David Forsyth
%{\tt\small daf@illinois.edu}
}
\institute{Z. Liao \at
              College of Computer Science, Zhejiang University, Hangzhou; AND --\\
              Department of Computer Science, University of Illinois at Urbana-Champaign, Champaign\\
              \email{zliao@zju.edu.cn}           %  \\
              %\emph{Present address:} of F. Author  %  if needed
           \and
           K. Karsch \at
           %Lumenous Corporation, San Francisco\\
           Department of Computer Science, University of Illinois at Urbana-Champaign, Champaign\\
           \email{karsch1@illinois.edu}
           \and
           H. Zhang \at
           College of Computer Science, Zhejiang University, Hangzhou \\
           %Department of Computer Science, University of California, San Diego\\
           \email{zhanghongyi@zju.edu.cn}
           \and
           D. A. Forsyth \at
           Department of Computer Science, University of Illinois at Urbana-Champaign, Champaign\\
           \email{daf@illinois.edu}
}

\maketitle
%\thispagestyle{empty}

%%%%%%%%% ABSTRACT
\vspace{-3mm}
\begin{abstract}
\RB{
We present an object relighting system that allows an artist to select an object from an image and insert it into a target scene. Through simple interactions, the system can adjust illumination on the inserted object so that it appears naturally in the scene. To support image-based relighting, we build object model from the image, and propose a \emph{perceptually-inspired} approximate shading model for the relighting. It decomposes the shading field into (a) a rough shape term that can be reshaded, (b) a parametric shading detail that encodes missing features from the first term, and (c) a geometric detail term that captures fine-scale material properties. With this decomposition, the shading model combines 3D rendering and image-based composition and allows more flexible compositing than image-based methods. Quantitative evaluation and a set of user studies suggest our method is a promising alternative to existing methods of object insertion.
}
%We propose a perceptually-inspired approximate shading model for image-based object modeling and insertion. The model is a hybrid of 3D rendering and image-based composition. It circumvents the difficulty of physically accurate shape reconstruction, and allows for more flexible composition than image-based methods. The model decomposes the shading field into (a) a rough shape term that can be reshaded, (b) a parametric shading detail that encodes missing features from the first term, and (c) a geometric detail term that captures fine-scale material properties.  With this object model, we build an object relighting system that allows an artist to select an object from an image and insert it into a 3D scene. Through simple interactions, the system can adjust illumination on the inserted object so that it appears more naturally in the scene. Quantitative evaluation and user studies suggest our method is a promising alternative to existing methods of object insertion.
\end{abstract}

%%%%%%%%%%%%%%%%%%%%%%%%%%%%%%%%%%%%%%%%%%%%%%%%%%%%%%%%%%%%%%%%%%%%%%
\section{Introduction}
\begin{figure}%[!t]
\begin{minipage}{\textwidth}
\centering{
\includegraphics[width=0.49\linewidth]{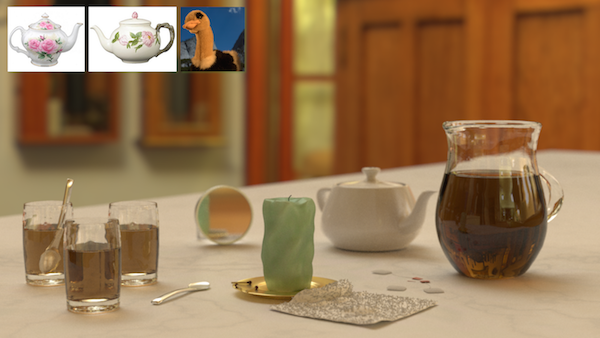} %\hspace{2mm}
\includegraphics[width=0.49\linewidth]{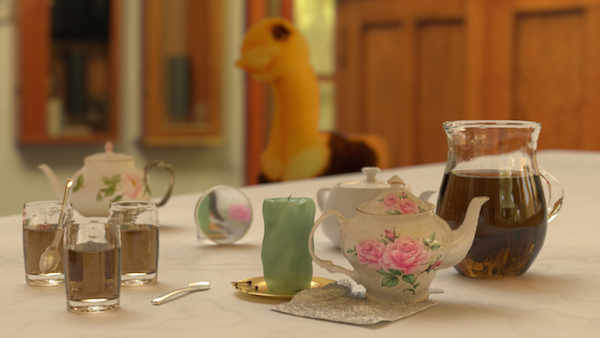}
}
\caption{Given an existing 3D scene (left), our method builds approximate object models from image fragments (two teapots and an ostrich) and inserts them in the scene (right).
Lighting and shadowing of the inserted objects appear consistent with the rest of the scene. Our method also captures complex reflection (front teapot), refraction (back teapot) and depth-of-field of effects (ostrich). None of these rendering effects can be achieved by image-based editing tools, while our hybrid approach avoids the cost of accurate 3D modeling. \emph{3D scene modeling credit to: Jason Clarke}.}
\label{fig:3Dscenes}
\end{minipage}
\end{figure}

This paper describes a system that makes it possible to insert objects from a source image to a target image.  This system can correct the shading of both the source object and the target image to make it look as though the object was originally part of the target image (Figure~\ref{fig:3Dscenes}).  Our system extends the state of the art, because we do not require any 3D information to be provided for the object that is being inserted -- it is represented as an image fragment, and user experience of authoring images is greatly simplified --- a user just selects an object in one image, and places it in another.

There is a significant literature dealing with methods to insert computer generated objects into real images, which we review below. The key problem here is to adjust the illumination of the object to be consistent with that of the scene. But rendering the object is straightforward, because it is a CG object. The system has a representation of geometry and material properties created by some modeling program.

There is also a significant literature on methods to compose images from fragments of other images. The key problem here is to avoid inconsistent illumination fields, which seem to be a strong cue that the target image isn't real. This is handled by managing the choice of fragments that are composed. Ideally, a system would \emph{relight} fragments to avoid inconsistency. The natural strategy is to try and recover a comprehensive geometric and material representation for the object from the image fragment. Doing so is one of the main open challenges in computer vision; as we show, current methods produce marginal relighting results.

The core observation is that one does not need to relight an image with fastidious physical accuracy to fool people. This view is supported both by the perception literature, reviewed briefly below, and by our user studies. Instead, we build a model with four separate components that can be estimated effectively: a reflectance term, a rough base shape, a parametric shading residual layer and a geometric detail layer. The reflectance term encodes object material; the base shape encodes coarse-scale shading by reshading, and the two detail layers encode higher-frequency components left out by the coarse shape. The base shape and the detail layers together encode the shading of an object. The model and the estimation procedures are described in section~\ref{sec:createmodel}; and the resulting relighting system in section~\ref{sec:relighting}.

The model is not intended to be physically accurate, but does give better re-rendering mean-squared error (MSE) than the state-of-the-art shape reconstruction method SIRFS~\cite{Barron:2012B} on the MIT intrinsic image dataset and a new dataset. The model shines in user studies of performance.  A set of extensive user studies shows that subjects preferred our results over that of SIRFS by a margin of 20\%, and over that of Karsch et al.~\cite{Karsch:2011} by a margin of 14\%, indicating the model provides a promising alternative solution to existing methods of object insertion.

%%\iffalse
\section{Related work}\label{sec:related}
%
% cases: cg->real, real->cg, real->real
%

\Ra{
To insert an object into a scene, a method must (a) determine the appearance of the object, when in the scene and (b) determine the effect of the object on  the scene's appearance. Different scenarios compel different approaches to these problems.  For the cases of interest to us, each problem is attacked by recovering some form of geometric and photometric model of object and scene, then predicting appearance with these models.  Relevant methods are linked by using the same compositing approach.

{\bf Compositing:} In each of the cases of interest, we will have multiple estimates of most pixels, and must come up with a single image.
 Usually, there is $I_t$ (the target image of the scene the object must be inserted into), $I_e$ (a rendered image of a model of the scene without the inserted object), and $I_r$ (a rendered image of a model of the scene with an inserted object).   Generally, we wish to composite these estimates to use  the most reliable estimate at each pixel.    We assume we have an object matte $M$ (which is 0 at pixels where the object is absent, and $(0,1]$ otherwise).  Debevec et al. describe a now-standard method that preserves the original image as much as possible~\cite{Debevec:1998:RSO}
In this method, the final composite image $C_{\text{final}}$ is obtained by:
\begin{equation}
\label{eq:composite}
C_{\text{final}} = M \odot I_r + (1-M) \odot (I_t + I_r - I_e).
\end{equation}

% Core problems:
%   - estimate illumination field object encounters
%   - estimate geometric and photometric object parameters
%   - estimate effects of object on scene shading field (cast shadows, etc.)
%       - by estimating scene geometric and photometric parameters

Methods differ by how the object (resp. scene) model is estimated.  Great simplifications are available if
either scene or object is CG, or either scene or object is readily accessible (i.e. one can obtain many images, under a wide range of conditions; contrast  an object or scene depicted in a legacy image).
Symmetry means we can consider only cases where the scene is real (as in, not CG).  We ignore the case where both scene and object
are CG, which is rendering.

{\bf CG object into accessible scene:} If the object is CG, then its geometric and photometric parameters are known.  Strategies then differ depending on what is known about the scene.  In one case of great practical importance, one has detailed access to the real scene into which a CG object must be  inserted.   Debevec {\em et al.} have shown methods that use reference objects to estimate the illumination environment at various points in the scene~\cite{Debevec:1998:RSO}.  From this estimate, one can relight the object.  If the scene illumination field needs to be corrected to account for the object's effects, then one uses a geometric and photometric model of the scene.

{\bf CG object into legacy image:} If the scene is a legacy image, there are significant difficulties.  One must estimate sufficient geometric and photometric information from the image to be able to compute the object shading and any effects on the scene illumination.  The legacy image is $I_t$.  Karsch et al. use a simple geometric model of a room as a box, estimate albedo using standard Color Retinex algorithm~\cite{Grosse:2009}, and estimate luminaires by looking for bright image patches~\cite{Karsch:2011}.   This model is corrected by user interaction.  The luminaire parameters are then adjusted to make the rendered scene similar to $I_t$ in $L_2$ norm.  Finally, the model is rendered to produce $I_e$.   The object is then inserted using a geometric modeler.  A simple renderer produces $M$ from this information; $I_r$ is obtained with a physically-based renderer.  These images are composited as above.

The original method requires user interactions.  A later paper by Karsch et al. describes an entirely automatic method for recovering a scene model~\cite{Karsch:2014}.  Scene geometry is estimated using the depth from single image method of~\cite{Karsch:depth2012}.  Visible luminaries are obtained by looking for very bright patches.  The effect of out of view luminaires is estimated using a matcher.  Albedo is again estimated using standard methods~\cite{Grosse:2009}.   Once the scene model has been recovered, the pipeline is the same as that in~\cite{Karsch:2011}.

{\bf Accessible real objects} reduce to the case of CG objects.  One uses a variety of standard methods to build geometric and photometric models of the object, and proceeds as above.  Modeling methods are reviewed in~\cite{Hartley2004,photometricstereo:2008,Furukawa2015}.  There is a highly developed literature for the very important case of  faces, recently reviewed in~\cite{Ghosh:2011:MFC,Fyffe:2013:DHF,Fyffe:facetopology2017}.  An alternative, which we believe has not been explored in the literature, would be to recover
an \emph{illumination cone} for the fragment from multiple images.  An illumination cone is
a representation of all images of an object in a fixed configuration, as lighting changes.
This cone is known to lie close to a low dimensional space~\cite{Basri:2003} -- a 9-Spherical Harmonics illumination can account for up to 98\% of shading variation, suggesting that quite low-dimensional image based reshading methods are available.   We use illumination cone methods in relighting objects, but have not explored the illumination cone as a modelling strategy.
%Standard methods to variation in appearance c

{\bf Objects from legacy images into other legacy images} are the primary topic of this paper.  Here one wishes to cut a fragment, representing an  object, out of one image and insert it into another, obtaining realistic results.  Early work focused on resolving blending issues, and relied on the artist's discretion not to insert an incompatible fragment into an image~\cite{Burt83amultiresolution,Perez:2003:poisson,Agarwala:2004:montage}.   The artist's work can be simplified by building a dictionary of fragments, organized by illumination estimates.  This means that, for a particular scene, the artist can see and choose from compatible fragments\cite{Lalonde:2007:clipart,Chen:2009:sketch2photo,Liao:CVPR15}.    However, there is no means to relight a fragment or correct the shading of a scene.

One natural strategy would be to recover a full geometric and photometric model of the object from the legacy image.  This strategy is not available, because doing so is one of the major open problems in computer vision.  Current methods are still unable to recover accurate shape from a single image, even inferring satisfactory approximate shape is difficult. Methods that recover \emph{shape from shading} (SfS) are unstable and inaccurate, particularly in the absence of reliable albedo and illumination~\cite{Zhang:1999:survey,Durou:2008:survey}. Barron and Malik have shown
significant improvements from jointly estimating shape, illumination and albedo~\cite{Barron:2012B}.  We compare to that method here, though the method is not intended to cope with objects that have complex geometric mesostructure.  An alternative would be to recover \emph{shape from contour}, to recover a surface that is smooth and meets normal constraints along the contour~\cite{Johnston:2002:lumo,Prasad:2006:SVR}.  These methods yield surface reconstructions that lack any detail.  Our method revolves around correcting shading predictions made by a variant shape-from-contour method.

%%{\bf Recovering material estimates:}  We use the color Retinex algorithm to estimate the albedo of surfaces. This is a variant of the original Retinex~\cite{Land:1971}, which is known to work well~\cite{Grosse:2009}.

%{\bf Recovering illumination estimates:}

{\bf Human tolerance of shading inaccuracy:}  Evidence shows that human visual system can tolerate certain degrees of shading inaccuracy.  For example, observers find it hard to spot inconsistent shadow directions in a single image~\cite{Ostrovsky:2005} as long as gross shading is correct.   Highlights are important material cues for humans~\cite{Beck:1981:HPG}, but observers are not perturbed if the highlight is somewhat in the wrong place (see~\cite{Berzh:2005:HGP}, experiment 3). The \emph{alternative physics} theory~\cite{Cavanagh:2005} argues that the brain employs a set of rules that are convenient, but not strictly physical, when interpreting a scene from retina image. When these rules are violated, a perception alarm is fired, or recognition is negatively affected~\cite{Tarr:1999}. Otherwise, the scene ``looks right''.    This means that humans may tolerate a considerable degree of estimation error, as long as it is of the right kind.  Our results -- which are clearly neither canonical nor physical, but are effective at fooling human viewers -- support this notion.

We note a poorly understood asymmetry between scene and object here, caused by the way in which methods are used.  Typically, the object that will be inserted is a crucial part of the target image, and will command much visual attention from the observer.  This means that humans may react differently to errors in depiction of object and scene.  We are not aware of guidelines in the literature as to what is tolerable here.
It is usual to build less scrupulous scene models than object models, and doing so seems not to present difficulties (eg~\cite{Karsch:2011,Karsch:2014,Debevec:1998:RSO}).
}
%%\fi
%%%%%%%%%%%%%%%%%%%%%%%%%%%%%%%%%%%%%%%%%%%%%%%%%%%%%%%%%%%%%%%%%%%%%%
\begin{figure*}%[t]
  \centering
  \includegraphics[trim=30 50 30 70, clip, width=\linewidth]{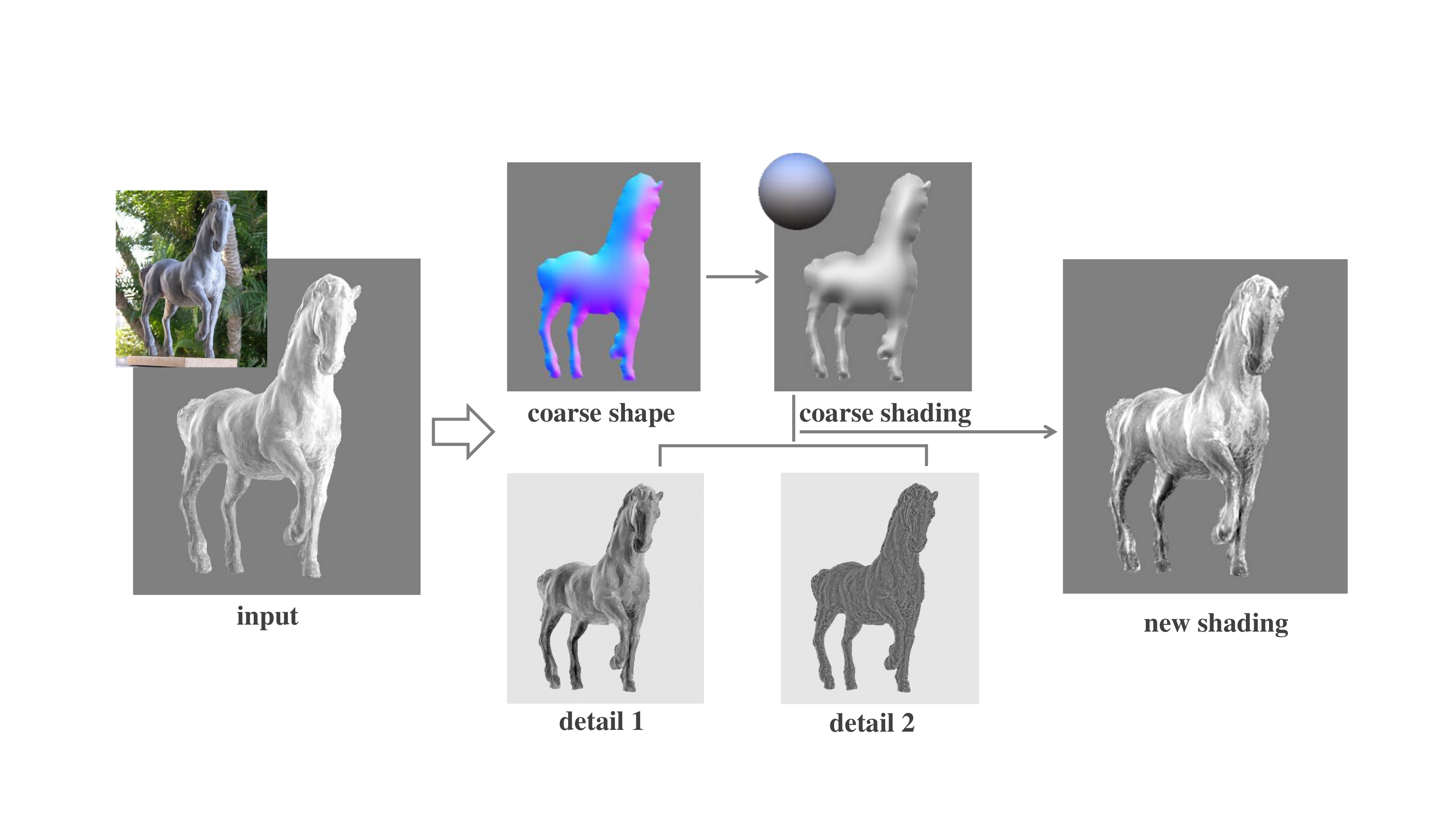}\\
  \caption{Given an image fragment with albedo-shading decomposition (left, albedo is omitted in this illustration), we build a coarse shape from contour; we then derive two shading detail layers (\emph{detail 1} and \emph{detail 2}) from the shading image and the coarse shape (middle). Our approximate shading model reshades the object under new illumination and produces a new shading on the right. Notice the change in both gross shading (new lighting from above) and the surface detail. The detail images are rescaled for visualization.
  }\vspace{-2mm}
  \label{fig:illustration}
\end{figure*}

\section{The shading model}\label{sec:createmodel}

We have an image fragment that is shaded by some illumination field.  The fragment represents an object which might have complex surface material.  There may be gloss, specular or mesostructural effects.  We wish to adjust the shading over the fragment to look as though the illumination field has changed, while preserving the apparent material effects.  We expect that errors in gross shading effects -- for example, which side of an object illuminated from the side is bright -- are likely to be apparent to people.  We expect that inconsistencies between gross shading and small shadows (say, from mesostructural bumps) may be difficult to spot.  We expect to be able to recover only a very simple geometric and photometric model of the object from the fragment.

These observations motivate our approximate shading model.  We decompose shading into 3 components -- a smooth component captured by shading a coarse geometric shape model $Z$ (the ``coarse shape''), and two shading detail layers: a parametric residual layer $S_p$ and a geometric detail layer $S_g$. The relighting of the object is given by the smooth shading plus a weighted combination of the two detail layers:
\begin{equation}\label{eq:model}
    S(Z, S_p, S_g; L, w) = \mbox{\emph{shade}}(Z, L) + w_pS_p + w_gS_g
\end{equation}
where $L$ is an illumination field,  $w_p$ and $w_g$ are scalar weights, and
%$S_p$  and $S_g$ are referred as \emph{detail 1} and \emph{detail 2} throughout the paper. %$L$ can be a point source or an area light when the renderer ($shade(h,L)$) is an off-the-shelf physically-based renderer, or a Spherical Harmonics representation of environment light with the corresponding analytical shading procedure
$\mbox{\emph{shade}} (Z,L)$ can be any shading (rendering) procedure that works for the illumination and scene representation.
%, for example, it could be an off-the-shelf physically-based renderer where $L$ can be point sources or area light, or it could take an efficient analytical form for the Spherical Harmonics light representation.

%*** SORT OUT ALBEDO ***

By shading the coarse shape, we can obtain a smooth shading that captures directional long-scale illumination effects.  These make it look as though the overall illumination direction has changed.   The two detail layers account for appearance complexity of the object. They encode mid and high frequencies of the appearance that are not successfully predicted by the albedo and shape alone.
We see a loose correspondence between the mid frequency layer ($S_p$) and small shape features (silhouettes, creases and folds, etc.), and between the high frequency layer ($S_g$) and material properties. Separating the two allows an artist to adjust each weight ($w_p$ and $w_g$ in equation~\ref{eq:model}) independently.

% \section{Creating object model from image}\label{sec:createmodel}
% Our shading model has four components for an object.  We first compute a coarse 3D shape estimate, then compute three maps: the albedo, a parametric shading residual, and a geometric detail layer. We refer to the ``coarse shading'' (encoded by the shape), the ``parametric shading residual'' and the ``geometric detail'' as the three shading components.

\subsection{Estimating coarse shape}\label{subsec:coarseshading}

We need a shape representation capable of capturing gross shading effects.  For example, a vertical cylinder with light from the left will be light on left, dark on right.  Moving the light to the right will cause it to become dark on left, light on right. We also want our reconstruction to be consistent with generic view and generic lighting assumptions. This implies that the outline should not shift too much if the view shifts.  It also implies relighting the shape with a directional light should not produce large or curious cast shadows.  In turn, there should not be large bumps in the shape that are concealed by the view direction (i.e. that extend toward the viewer).  Large bumps in shape concealed by the view direction are characteristic of current shape from shading methods (Fig.~\ref{fig:cube}), but can create large cast shadows in the scene.

%We also want a representation that supports generic view direction. Traditional shape from shading is prone to produce depth maps with large errors (spikes, large bumps, etc.), which can make deceivingly good images when reshaded in exactly the same view direction as in the original image, but cause serious occlusion and expose the error when the view direction is slightly changed (\cite{Karsch:CVPR13} Fig.3). To create shape that supports generic views, we use a simple shape from contour method with stable outline and no large bumps.
%Generic view direction for shape representation is an important problem in computer vision. It requires a shape representation to be safely viewed from non-coincidental directions. Traditional shape from shading is prone to produce depth map with large errors (spikes, extrusions, etc.), which can make deceivingly good images when reshaded in exactly the same view direction as in the original image, but causes serious occlusion and exposes the error when the view direction is slightly changed (\cite{Karsch:CVPR13} Fig.3). The smooth normal prior, flat base, and crease along boundary in our shape model, though appear extremely simple, are a set of principles for the support of generic view directions.

We use a simple shape from contour (SFC) method with a modification to ensure a stable outline (Fig.~\ref{fig:illustration}).
First, we create a normal field by constraining normals on the object boundary to be perpendicular to the view direction, and interpolate them from the boundary to the interior region, similar to Johnston's Lumo technique~\cite{Johnston:2002:lumo}. Let $N$ be the normal field, $\Omega$ and $\partial\Omega$ be the set of pixels in the object and on boundary, respectively. We compute $N$ by the following optimization:
\begin{equation}\label{eq:normal}
\begin{aligned}
& \underset{N}{\text{min}} & & \sum_{\Omega} {||\nabla {\bf N}||^2 + (||{\bf N}||^2-1)^2} \\
& \text{subject to} & & N_{z}^{i} = 0 %\hspace{2mm} \text{and} \hspace{2mm} N^{i}\cdot N_{\bot}^{i} = 0,
\hspace{3mm} \forall i \in \partial\Omega
\end{aligned}
\end{equation}

We then reconstruct an approximate height field $h$ from the normal by minimizing:
\begin{equation}
\sum_{\Omega}{ ||(\frac{\partial Z}{\partial x} + \frac{N_{x}}{max(\epsilon, N_z)})||^2 +
||(\frac{\partial Z}{\partial y} + \frac{N_{y}}{max(\epsilon, N_z)})||^2}
\end{equation}
subject to $Z_{i}=0$ for boundary pixels (stable outline).
The threshold  avoids numerical issues near the boundary for exact reconstruction (Wu et al.~\cite{Wu:2008:SFS}); and forces the reconstructed object to have a crease along its boundary. This crease is very useful for the support of generic view direction, as it allows slight change of view direction without exposing the back of the object and causing self-occlusion. The reconstructed height field is then flipped to make a symmetric full 3D shape.
Our model is a simple shape model in the spirit of puffballs~\cite{Adelson:puffball}, but in contrast has a crease along its boundary.
%*** NEED A COMPARISON TO PUFFBALLS

%\begin{figure}[t]
%\centering
%%\begin{minipage}{0.95\textwidth}
%\subfigure[normal]{\includegraphics[height=1in]{../images/shape/normalmap3}}\hspace{.4mm}%\hfill
%\subfigure[3D shape]{\includegraphics[height=1in]{../images/shape/mesh1new3}}\hspace{.4mm}%\hfill
%\subfigure[view2]{\includegraphics[height=1in]{../images/shape/mesh2new3}}\hspace{.4mm}%\hfill
%\subfigure[view3]{\includegraphics[height=1in]{../images/shape/mesh3new3}}
%%\end{minipage}
%\caption{Shape reconstruction. Our reconstruction is simple but robust to large errors that may occur in state-of-the-art SfS algorithm and supports the generic view assumption (Fig.~\ref{fig:cube}). %This benefit is key to our goal of image fragment insertion.}\label{fig:shape}
%\vspace{0mm}
%\end{figure}

\subsection{Parametric shading residual}\label{subsec:parresidual}
The coarse shape can recover gross changes in shading caused by lighting. However, it cannot represent finer detail. We use shading detail maps to represent this detail. We define the shading detail maps as a representation of the residual incurred by fitting the object shading estimate with some model. We use two shading details in our model: \emph{parametric shading residual} that encodes object level features (silhouettes, crease and folds, etc.), and \emph{geometric detail} that encodes fine scale material effects.

First, we use a standard color Retinex algorithm~\cite{Grosse:2009} to get an initial albedo $A$ and shading $S$ estimate from the input image: $I = A \cdot S$. We then use a parametric illumination model $L(\theta)$ to shade the coarse shape and compute the residual by solving:
\begin{equation}\label{eq:argminL}
\hat{\theta}: \underset{\theta}{\operatorname{argmin}} \sum{||S - \mbox{\emph{shade}}(Z, L(\theta))||^2}
\end{equation}
The optimized illumination $\hat{\theta}$ is substituted to obtain the parametric shading residual:
\begin{equation}\label{eq:computePS}
S_p = S- \mbox{\emph{shade}}(Z, L(\hat{\theta})).
\end{equation}
Many parametric illuminations are possible (i.e., spherical harmonics). We used a mixture of 5 point sources in all of our experiments, the parametrization of the lights are position and intensity of each point source, forming a 20-dimensional representation.

\begin{figure}[h]
\centering
\includegraphics[trim=0 0 0 0, clip, width=0.198\linewidth]{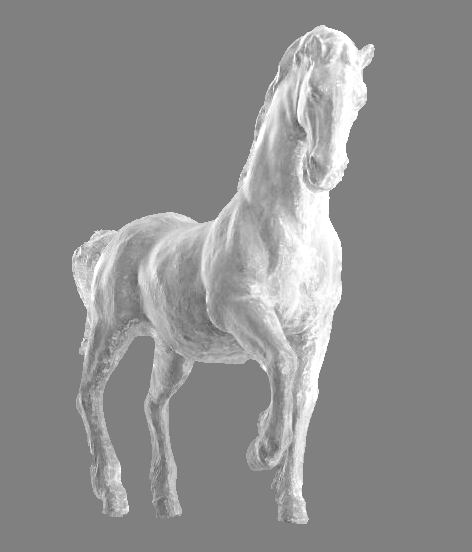}\hfill
\includegraphics[trim=0 0 0 0, clip, width=0.198\linewidth]{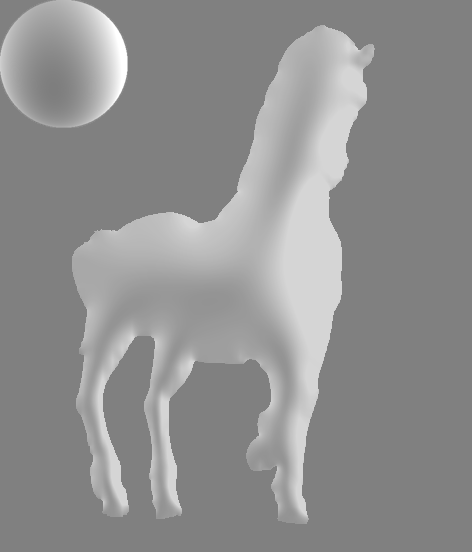}\hfill
\includegraphics[trim=0 0 0 0, clip, width=0.198\linewidth]{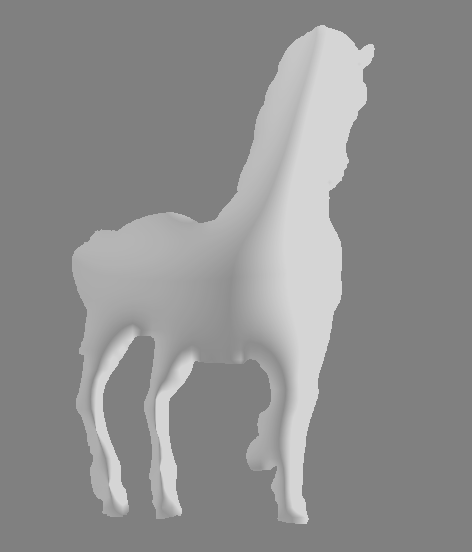}\hfill
\includegraphics[trim=0 0 0 0, clip, width=0.198\linewidth]{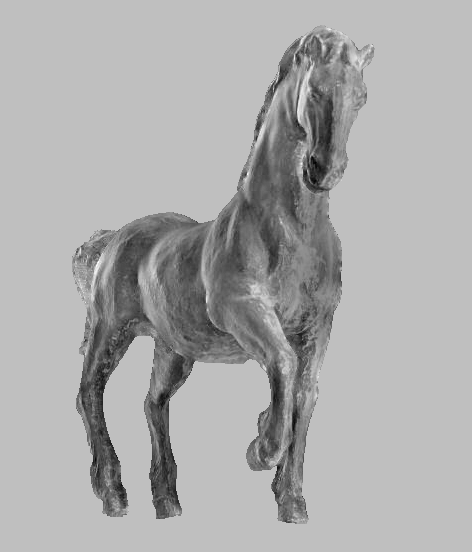}\hfill
\includegraphics[trim=0 0 0 0, clip, width=0.198\linewidth]{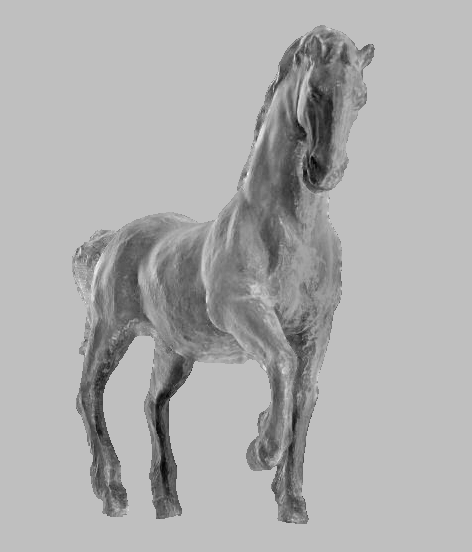}\\\vspace{.5mm}
\includegraphics[trim=0 0 0 0, clip, width=0.198\linewidth]{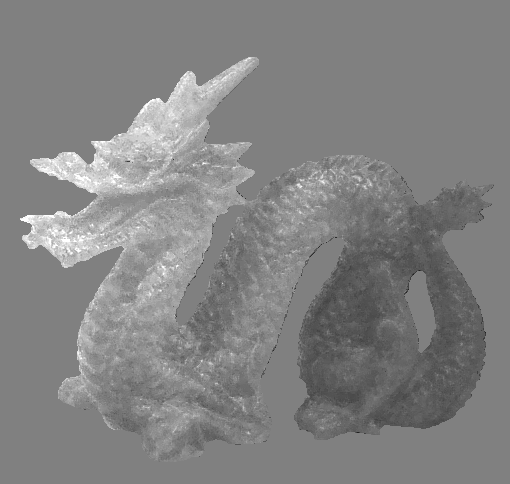}\hfill
\includegraphics[trim=0 0 0 0, clip, width=0.198\linewidth]{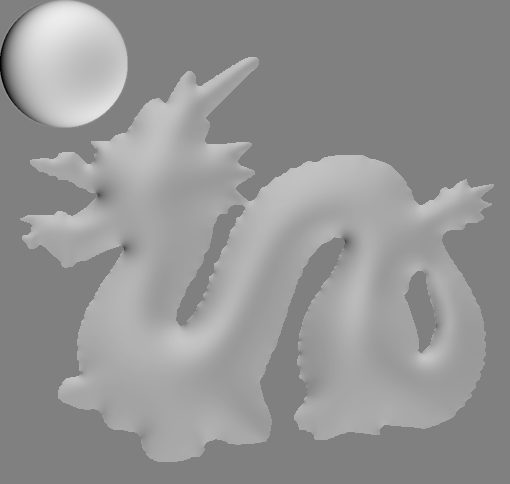}\hfill
\includegraphics[trim=0 0 0 0, clip, width=0.198\linewidth]{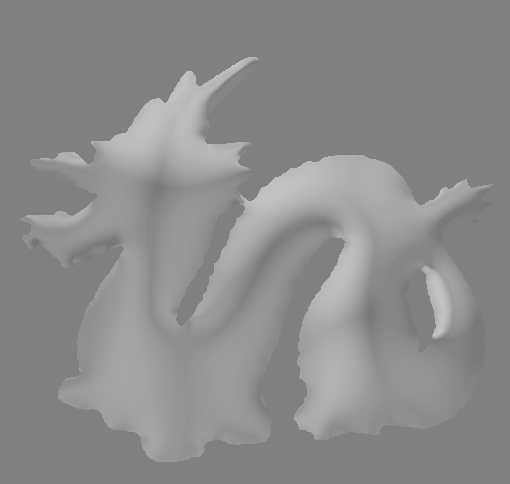}\hfill
\includegraphics[trim=0 0 0 0, clip, width=0.198\linewidth]{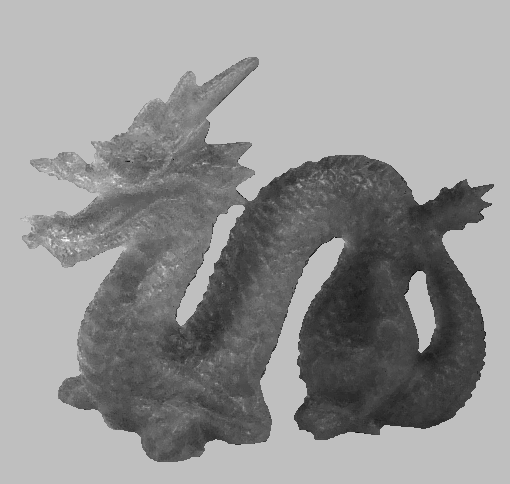}\hfill
\includegraphics[trim=0 0 0 0, clip, width=0.198\linewidth]{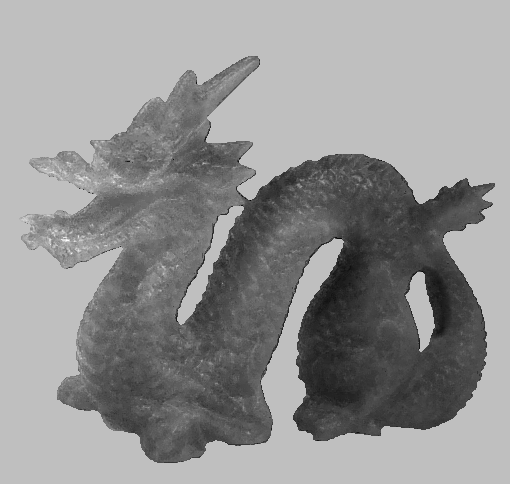}\\
\hspace{10mm}{\bf (a)}\hfill{\bf (b)}\hfill{\bf (c)}\hfill{\bf (d)}\hfill{\bf (e)}\hspace{10mm}
\caption{\Ra{Parameteric shading residual inferred from two different lighting representations. (a): input shading; (b) shading of the inferred spherical harmonics light on the coarse shape (visualization of inferred light is shown in upper left); (c) shading of the inferred point lights on the coarse shape (described in the paper). Note the point lights capture sharper shading variations; (d) parametric shading residual by subtracting (b) from (a); (e) parametric shading residual by subtracting (c) from (a).}}
\label{fig:SHlight}
\end{figure}
\Ra{An alternative is to use spherical harmonic lighting models. A comparison is made in Figure~\ref{fig:SHlight}, which shows parametrically fitted shading image and the derived shading residual using a 9-coefficient spherical harmonics representation versus the point lights. Note the shading image produced by a spherical harmonic model is smoother than that produced by a point source model.
Our coarse shape has relatively smooth normals, and SH lighting models do not account for occlusion well. In contrast, point source shading falls off faster as a normal swings, so our point source model can produce shading that changes more abruptly (e.g. the dragon in the third column of Figure~\ref{fig:SHlight}) In all of our experiments, we use our point source model to derive the parametric shading residual.}
%This is because our coarse shape is smooth, and SH lighting only keeps a few low frequency lighting patterns. The point light sources, instead, can make up for the smoothness nature of our coarse shape and produce shading that changes more abruptly (e.g. the dragon in third column of Figure~\ref{fig:SHlight}). In all of our experiments, we use the aforementioned point sources to derive parametric shading residual.}

\RB{Note that the use of five point light sources is out of empirical design consideration. The five point sources are initialized to large cover light from the five major directions (frontal and four sides). A larger number of point sources could be used, or other types of light forms, such as area light or the spherical harmonics. What we demonstrate in the comparison in Figure~\ref{fig:SHlight} -- in the case of the statue horse and the statue dragon --  is that the use of spherical harmonics and the point sources fit the input shading slightly differently, and more important, the resultant parametric shading detail still look characteristically very similar (Figure 3 (d) and (e)).
}

\begin{figure}[t]
\centering
\includegraphics[trim=50 20 50 30, clip, width=\linewidth]{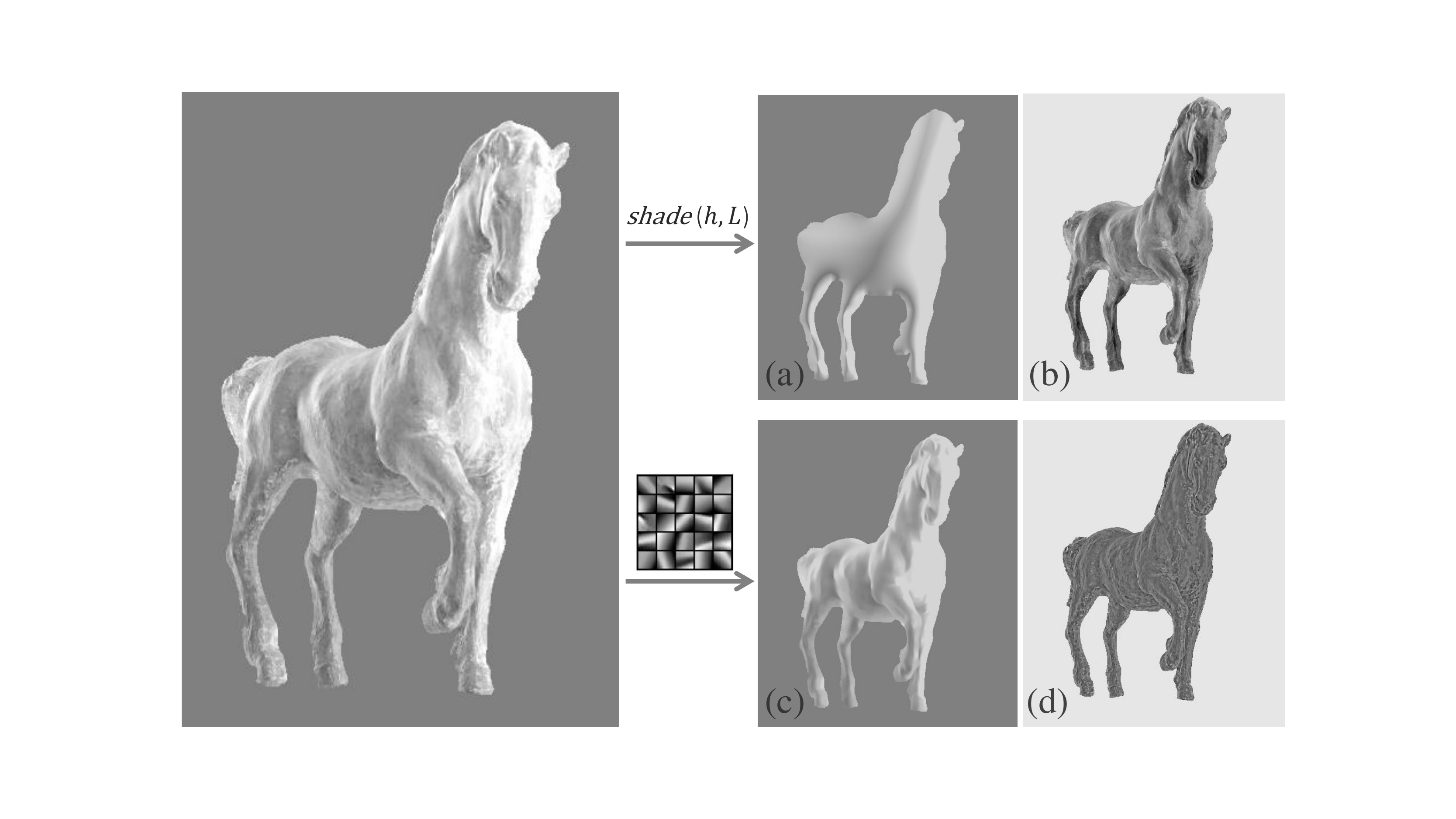}
\caption{Illustration of shading detail computation. Given the shading image (left), the upper right row shows the parametric fitting procedure that computes the best fit shading (a) from the shape and the parametric shading residual (b); the bottom right row shows the non-parametric patch-based filtering procedure that computes the filtered shading image (c) and the residual known as geometric detail (d).}
\label{fig:detailmaps}
\vspace{-3mm}
\end{figure}

Figure~\ref{fig:detailmaps} upper right row shows an example of the best fit coarse shading and the resultant parametric shading detail. Note that the directional shading is effectively removed, leaving shading cues of object level features. The bottom right row shows the geometric detail extraction pipeline.

\subsection{Geometric detail}\label{subsec:geometricdetail}
The parametric shading residual is computed by a \emph{global} shape and illumination parametrization, and contains all the shading detail missed by the shape. Now we wish to compute another layer that contains only fine-scale detail. \Ra{We use a technique from Liao et al.~\cite{Liao:2013:geometricdetail} that extracts fine-scale geometric detail. The algorithm reconstructs local image patches with an L1 regularized patch dictionary learned from data (called a \emph{local} patch-based non-parametric filtering) and computes the reconstruction residual as the geometric detail.} The resultant geometric detail represents high frequency shading signal caused by local surface geometry like bumps and grooves and is insensitive to gross shading and higher-level object features. See the characteristics of the geometric detail we obtained from the horse image (bottom row of Fig.~\ref{fig:detailmaps}) and compare it with the parametric shading residual term.

The filtering procedure uses a set of shading patches learned from smooth shading images to reconstruct an input shading image. %Because geometric detail signals are poorly encoded by the smooth shading dictionary, they are effectively left out. %See Liao et al.~\cite{Liao:2013:geometricdetail} for more details.
\Ra{In the experiment we use the same empirical parameter settings used in \cite{Liao:2013:geometricdetail}: dictionary size  of 500 patches and patch size $12 \times 12$. The two hyper-parameters should correlate with one another and should vary subject to input image size (image size in our experiments are around $600 \times 600$).
}

\vspace{2mm}
\noindent{\bf How many detail layers are necessary?} \Ra{
We choose two shading detail layers for a balance of representational power and ease of editing. Having two detail layers allows user to adjust mid- (eg. muscle) and high (eg. wrinkle) frequency shading components separately, which by itself is a straightforward design choice. However, better solutions should exist. One possible idea is to extend the two-layer structure to a Laplacian pyramid like decomposition, in which a larger number of detail layers is defined, each taking up a particular band of the frequency domain. Another possible idea is to replace the current image-based shading detail representation with \emph{illumination independent} ``detail normal maps'', which would have interesting implication to the relighting effect. Yet another possibility is to manually define new shading detail maps in the current framework that capture other types of high frequency shading phenomenon.
}

%We choose two layers empirically for a balance of representational power and the ease of editing. On the one hand, it is reasonable to increase the number of detail layers. However, this would make the editing interface less easy to use (more buttons on the user interface in Fig.~\ref{fig:rendercomposite}). On the other hand, having two detail layers allows user to adjust mid (eg. muscle) and high (eg. wrinkle) frequency shadings separately, simulating physical shading changes in a more flexible way.

\vspace{0mm}
\begin{figure}[t]
\centering
\includegraphics[width=0.9\linewidth]{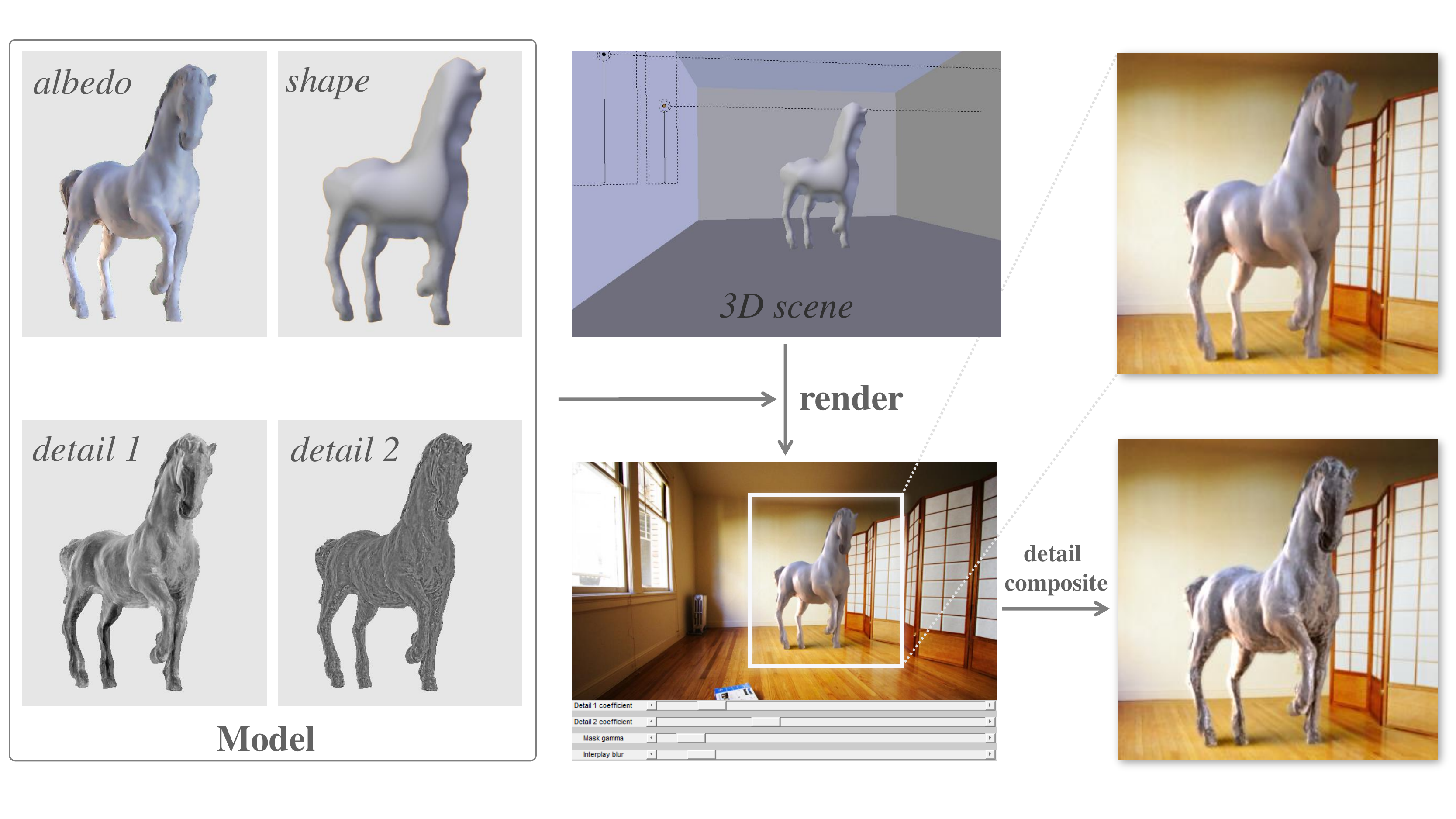}
\caption{Illustration of the relighting system. Given an object model (the horse), an artist places the model into a 3D scene, render it with a physically-based renderer, and then composite it with the detail layers to generate the final result (a close-up view in bottom right corner). Notice the difference of surface detail appearance on the horse before and after detail composition.}
\label{fig:rendercomposite}
\end{figure}
%%%%%%%%%%%%%%%%%%%%%%%%%%%%%%%%%%%%%%%%%%%%%%%%%%%%%%%
\section{The relighting system}\label{sec:relighting}

We now describe a system to take a fragment from an image and insert it into a new scene, relighting it and the scene as required. The system combines interactive scene modeling, physically-based rendering and image-based detail composition (Fig.~\ref{fig:rendercomposite}).

\subsection{Modeling and rendering}

We build a sparse mesh object from the height field computed by the method of section~\ref{subsec:coarseshading} (by pixel-grid triangulation and mesh simplification \cite{Xia:2009}). The target scene can be existing 3D scenes, or recovered from an image.  In the latter case, we use the approach of Karsch et al.~\cite{Karsch:2011} to recover a 3D model of the scene from image. \Ra{The method asks a user to specify indoor scene room boundaries, and then automatically recovers interior illumination parameters that minimize re-rendering error. The recovered illumination does a fairly good job in producing consistent shading on the inserted object; adjustment of the illumination intensity is sometimes needed in practice. Having had the model of the scene and the object,} the user then places the object model into the scene, adjust its scale and orientation, and make sure the view direction is roughly the same as that of the fragment in the original image. The model is then rendered with a physically-based renderer. Finally, the residual shading fields ($S_p$ and $S_w$) are composited into the rendered image. For all the results in the paper, we use Blender (http://blender.org) for modeling and LuxRender (http://luxrender.net) for physically-based rendering. All target scenes were constructed using the technique of Karsch et al.~\cite{Karsch:2011} technique if not otherwise stated.

To create the mesh object from the height field, we flip the mesh along the contour plane to create a closed 3D mesh. However, the flipped shape model is thin along the base and this can cause light leaks and/or skinny lateral shadows. We use a simple user-controllable \emph{back extrusion} procedure to handle such cases. Specifically, the back extrusion asks a user to manually select a distance to extrude the back side of the mesh (since it was flipped and symmetric) to ensure full contact of the object bottom with the supporting surface. The extruded back is eased in the camera's viewing direction to make sure it is invisible.

Our shape model assumes an orthographic camera, while most rendering systems use a perspective camera. This will cause texture distortion during texture mapping. Since we expect the focal point to be (a) far from the camera, and (b) largely frontal, we can use a simple \emph{easing} operation to avoid self-occlusion and restore the texture field: write  $p_m=(x, y, h)$ for a vertex on the model, $p_f=(x, y, 0)$ for the coordinate of the vertex in the image plane, $f$ for the focal point, we replace $p_m=(x, y, h)$ with $p_f+h*(f-p_f)/||(f-p_f)||$. If the camera is orthographic, there is no change in vertex position, and for cameras that are distant along the z-axis compared to the $x$ and $y$ axes, the shift is small.
%\Ra{Figure~\ref{fig:easing} shows the effect without and with the easing operation. Notice the stretching and displacement of the eye of the horse in the first re-rendered result (Middle).} 

\iffalse
\begin{figure*}[t!]
\centering
%\begin{subfigure}[]{}
%\centering
\includegraphics[height=35mm]{easing_original.PNG}\hspace{.5mm}
%\end{subfigure}
%\begin{subfigure}[]{}
\includegraphics[trim=120 0  180 0, clip, height=35mm]{easing_horse_noeasing}\hspace{.5mm}
%\end{subfigure}
%\begin{subfigure}[]{}
\includegraphics[trim=120 0  180 0, clip, height=35mm]{easing_horse_witheasing}
%\end{subfigure}
\caption{\Ra{The effect without and with the easing operation. Left: original image of the horse; Middle: re-rendering without easing; Right: re-rendering with easing. Notice the distortion of the horse in the first rendering result (without easing): texture near the contour is pushed towards the boundary, e.g. the right eye is moved upward, and the tail squeezed.}}
\label{fig:easing}
\end{figure*}

\fi

%We use a simple ``\emph{easing}'' method to avoid this problem. %\zicheng{Details are included in the supplemental material (``Easing and back extrusion'' section) of the conference version of this work (Liao et al.~\cite{Liao:CVPR15})}.

\vspace{10mm}
\begin{figure}[h]
\begin{minipage}{0.33\linewidth}
\centering{\bf Inset} \\
\includegraphics[width=\linewidth]{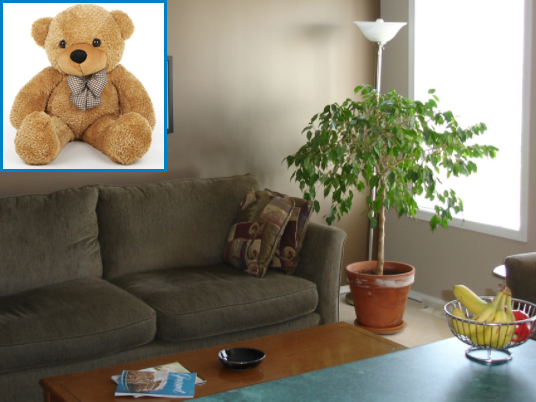}
\end{minipage}
\begin{minipage}{0.33\linewidth}
\centering{\bf B \& M} \\
\includegraphics[width=\linewidth]{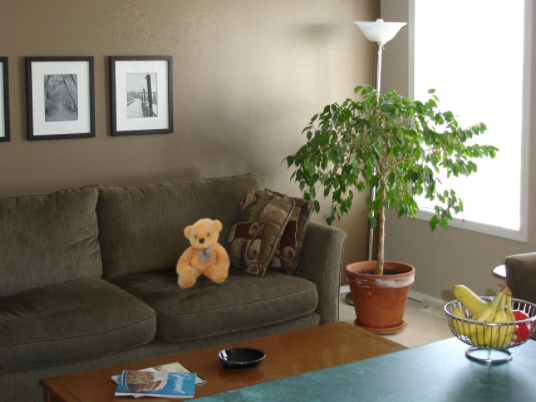}
\end{minipage}
\begin{minipage}{0.33\linewidth}
\centering{\bf Ours} \\
\includegraphics[width=\linewidth]{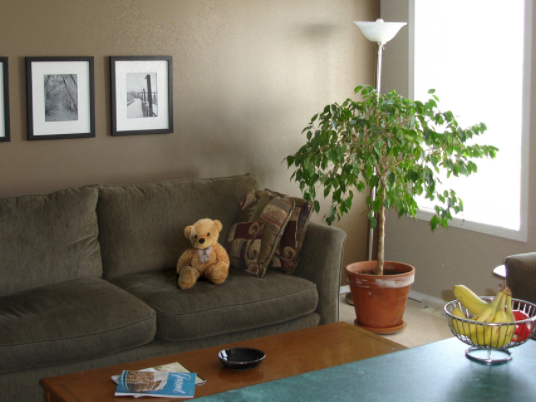}
\end{minipage}

\vspace{1mm}
\begin{minipage}{0.495\linewidth}
\includegraphics[trim=70 60  150 120, clip, width=\linewidth]{fruitroom-teddy_barron_cropped}
\end{minipage}
\hfill
\begin{minipage}{0.495\linewidth}
\includegraphics[trim=70 60  150 120, clip, width=\linewidth]{fruitroom-teddy_ours_cropped}
\end{minipage}
\caption{
Object relighting by our method (top right column), comparing with the method using shape estimates by Barron and Malik~\protect{\cite{Barron:2012B}} (top mid column, missing shadows due to its ``flat'' shape reconstruction).
Our relighting method adjusts the shading on the object for a variety of scenes with different illumination conditions.
Detail composition simulates complex surface geometry and materials properties that is difficult to achieve by physically-based modeling.
A close-up view of the result of Barron and Malik and ours is displayed in the bottom row. Best viewed in color at high-resolution.
}\vspace{0mm}
\label{fig:result1}
\end{figure}

\begin{figure}[!th]
\begin{minipage}{0.325\linewidth}
\centering{\bf No detail applied} \\
\includegraphics[width=\linewidth]{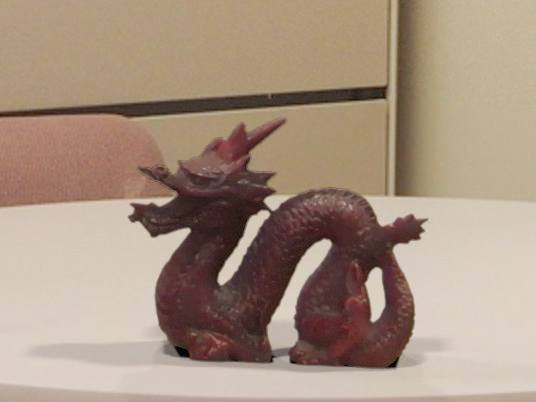}
\end{minipage}
\begin{minipage}{0.325\linewidth}
\centering{\bf Detail layer 1} \\
\includegraphics[width=\linewidth]{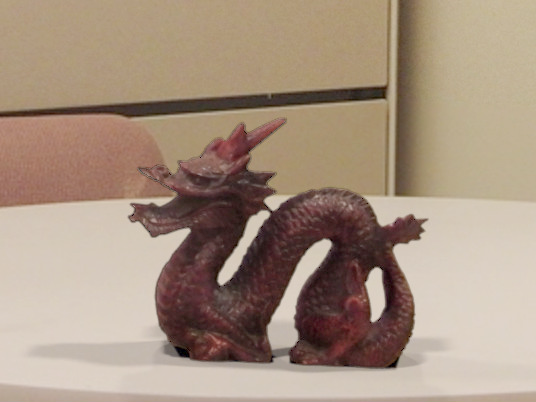}
\end{minipage}
\begin{minipage}{0.325\linewidth}
\centering{\bf Detail layer 1 + 2} \\
\includegraphics[width=\linewidth]{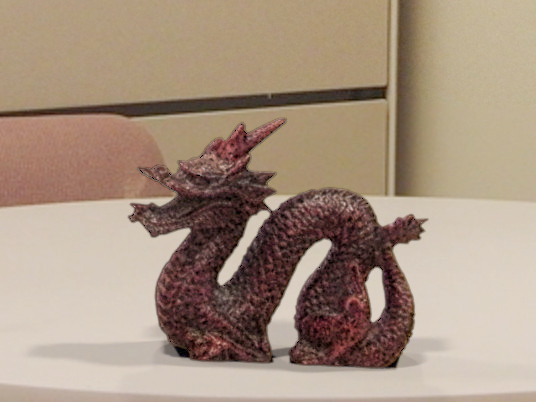}
\end{minipage}
\vspace{2mm}
\caption{
Detail composition. The left column displays relighting results with our coarse shape model and estimated albedo. The middle column displays results compositing with only the parametric shading residual. Notice how this component adds object level shading cues and improves visual realism. The right column are results compositing with both detail layers. Fine-scale surface detail is further enhanced (see the dragon). Best viewed in color at high-resolution.
}
\label{fig:detailedit}
\end{figure}

\vspace{-5mm}
\subsection{Compositing}\label{subsec:compositing}

We now composite the rendered scene with the two detail maps and the original scene to produce the final result (Fig.~\ref{fig:result1}).
First, we composite the two shading detail images with the shading field of the rendered image:
\begin{equation}
\label{eq:detail_composite}
C = A (S + w_pS_p + w_gS_g)
\end{equation}
where $S=I_r/A$ (equivalently denoted as $shade(h,L)$ in equation~\ref{eq:model}) is the shading field, $I_r$ is the rendered image. The weights $w_p$ and $w_g$ can be automatically determined by regression (section~\ref{subsec:rerender_error}) or manually adjusted by artist with a slider control (Fig.~\ref{fig:rendercomposite} detail composition). Compositing the two detail layers improves the visual realism of the rendered object (Fig.~\ref{fig:detailedit}). A controlled user study (section~\ref{subsec:userstudy}, task 1) showed that users consistently prefer composition results with more detail layers applied. \Ra{Note that the compositing procedure can tolerate certain degree of albedo-shading misclassification, because the albedo and the shading are composited back (equation~\ref{eq:detail_composite}). Misclassification of albedo and shading would, however, affect the detail manipulation results (when $w_p$ and $w_g$ are not equal to 1) in principle, i.e., color instead of shading signal gets magnified or smoothed.}

Second, we use standard techniques (e.g.~\cite{Debevec:1998:RSO,Karsch:2011}) to composite $C$ with the original image of the target scene. This produces the final result. %As in section~\ref{sec:relatedwork},
Write $I_t$ for the target image, $I_e$ for the empty rendered scene {\it without} the inserted object, $I_r$ for the rendered scene with the inserted object, and $M$ for the object matte (0 where no object is present, and $(0,1]$ otherwise), the final composite image $C_{\text{final}}$ is obtained by equation~\ref{eq:composite}.

\section{Evaluation}\label{evaluation}

Our assumption is that the approximate shading model can capture major effects of illumination change of an object in new environment and generate visually plausible image. To evaluate the performance, we compare our representation with state-of-the-art shape reconstructions by Barron and Malik~\cite{Barron:2012B} on a re-rendering metric (Section~\ref{subsec:rerender_error}). We then evaluated the affect of the different base shape acquisition methods on the relighting performance (Section~\ref{subsec:shapestudy}). We also conduct an extensive set of user study to evaluate the realism of our relighting results versus that of Barron and Malik~\cite{Barron:2012B}, Karsch et al.~\cite{Karsch:2011} and real scenes (Section~\ref{subsec:userstudy}). The evaluation results show that our representation is a promising alternative to the existing methods for object relighting.

\subsection{Re-rendering error}\label{subsec:rerender_error}
The re-rendering metric measures the error of relighting an estimated shape. On a canonical shape representation (a depth field), the metric is defined as
\begin{equation}\label{MSEmetric2}
\mbox{IMSE}_{rerender} = \frac{1}{n}||I - k \hat{A}\ \mbox{ReShade}(\hat{Z}, L)||^2
\end{equation}
where $\hat{A}$ and $\hat{Z}$ are estimated albedo and depth, $I$ is the re-rendering with the ground truth shape $Z^*$ and albedo $A^*$: $I = A^*\mbox{ReShade}(Z^*,L)$, $n$ is the number of pixels, $k$ is a scaling factor that minimizes the squared error.

With our model, write $S_c = shade(Z,L), S_p, S_g$ for the coarse shading, parametric shading detail and the geometric detail, respectively, and replace the corresponding part of Equation~\ref{eq:model} with $\mbox{ReShade}(S(L), w) = S_c + w_pS_p + w_gS_g$ for some choice of weight vector $w=(1,w_p,w_g)$. The re-rendering metric is:
\begin{equation}\label{eq:MSEmetric}
%\mbox{IMSE}^{'}_{rerender} = \frac{1}{n}||I - k \hat{\rho}\ \mbox{ReShade}(S(L),w)||^2
\mbox{IMSE}^{'}_{rerender} = \frac{1}{n}||I - k \hat{A}\ \mbox{ReShade}(S(L), w)||^2
\end{equation}
That is, rendering of canonical shape is replaced by our approximate shading model (Equation~\ref{eq:model}).

We offer three methods to select $w$. An {\em oracle} could determine the values by least square fitting that leads to best MSE. {\em Regression} could offer a value based on past experience. We learn a simple linear regression model to predict the weights from illumination. Lastly, an artist could {\em manually} choose the weights, as demonstrated in our relighting system (Sec.~\ref{subsec:compositing}).

\begin{table}[t]
\begin{center}
\begin{tabular}{l||cc|cc}
%\hline
Method &
\multicolumn{2}{c|}{\begin{tabular}[x]{@{}c@{}}``Natural'' Illumination\\ \scriptsize{(No strong shadows)}\end{tabular}} &
\multicolumn{2}{c}{\begin{tabular}[x]{@{}c@{}}Lab Illumination\\ \scriptsize{(With strong shadows)}\end{tabular}}\\
\hline\hline
{\em B + M}~\cite{Barron:2012B}     & \multicolumn{2}{c|}{0.0172}      & \multicolumn{2}{c}{0.0372} \\
\hline
 {\em Ours:} & {\em regress} & {\em oracle}  & {\em regress} & {\em oracle}  \\

{(a) default}                & 0.0358 & 0.0329   & 0.0641   & 0.0586 \\
{(b) $B+M S$}               & 0.0320 & 0.0274   & 0.0360   & 0.0341 \\
{(c) $GT S$}                 & 0.0243 & 0.0206   & 0.0240   & 0.0228 \\
%{(d) GT $S \& L$}           & 0.0149 & 0.0219 &  NA  & NA \\
%\hline
\end{tabular}
\end{center}
\caption{
Re-rendering error of our method compared to Barron \& Malik~\protect{\cite{Barron:2012B}}. Ours ``default'' uses models derived in the default setting. Ours ``B+M S'' uses Barron and Malik's shading estimation for detail derivation. Ours ``GT S'' uses the ground truth shading for detail derivation. Note our method performs less as good on a synthetic image dataset (the ``Natural'' Illumination dataset) where Barron and Malik produces very accurate shape estimates. On the real image dataset (the ``Lab'' Illumination dataset), our method with automatic weights can perform better than Barron and Malik.}
\label{tab:mse}
\end{table}

%{\bf \em What are the cases in table 1 (ie B+M S, GTS, default)?}

\vspace{3mm}
\noindent{\bf Experiment}
We run the evaluation on the augmented MIT Intrinsic image dataset~\cite{Barron:2012B}.
To generate the target images, we re-render each of the 20 object by 20 randomized monochrome ($9\times1)$ Spherical Harmonics illuminations, forming a $20\times20$ image set.
We then measure the re-rendering error of our model and Barron and Malik's reconstructions. For our method, we compare models built from the Natural Illumination dataset and Lab Illumination dataset separately. The models are built (a) in the default setting, (b) using Barron and Malik's shading estimation, and (c) using the ground truth shading. %and (d) using both ground truth shading and illumination.
See Table~\ref{tab:mse} for the results. To learn the linear regression model, we draw for each object 100 nearest neighbors in illumination space from the other 380 data points, and fit its weights by least square.

Table~\ref{tab:mse} displays our experiment result. The result shows that when the shape estimation of Barron and Malik is accurate (on the ``Natural'' Illumination dataset, a synthetic dataset by the same shading model used in their optimization), our approximate shading performs less well. This is reasonable because a perfect shape is supposed to produce zero error in the re-rendering metric. This is also acceptable because the dataset images are not real. On the real image set (the ``Lab'' Illumination dataset, taken in lab environment with strong shadows), the shape estimation of Barron and Malik becomes inaccurate, and our approximate shading model can produce lower error with both regressed weights and oracle setting. With better detail layers (when ground truth shading is used to derive them), our model achieves significantly lower errors.

%------------------------ study the affect of different shapes
\subsection{The effect of shape estimates}\label{subsec:shapestudy}

The model relies on the rough base shape to generate coarse-scale shading according to lighting environment. For
practical consideration we have chosen an internally-smooth SFC representation. This representation captures no
object-specific creases or folds, and can be quite different from the true depth.   In this section,  we investigate the
effect of alternative estimates of the rough base shape.

We consider three alternative shape representations: (a) \emph{Kinect} depth, (b) depth from \emph{SIRFS}, and (c) depth
from SIRFS-stereo.  These are fair baselines.  The Kinect is a standard consumer depth sensor.  The SIRFS and SIRFS-stereo algorithm are described in Barron and Malik~\cite{Barron:2012B}; the latter was used to reconstruct the ground-truth depth of the augmented MIT intrinsic image dataset using multiple differently illuminated images.

\begin{figure*}
  \includegraphics[width=\linewidth]{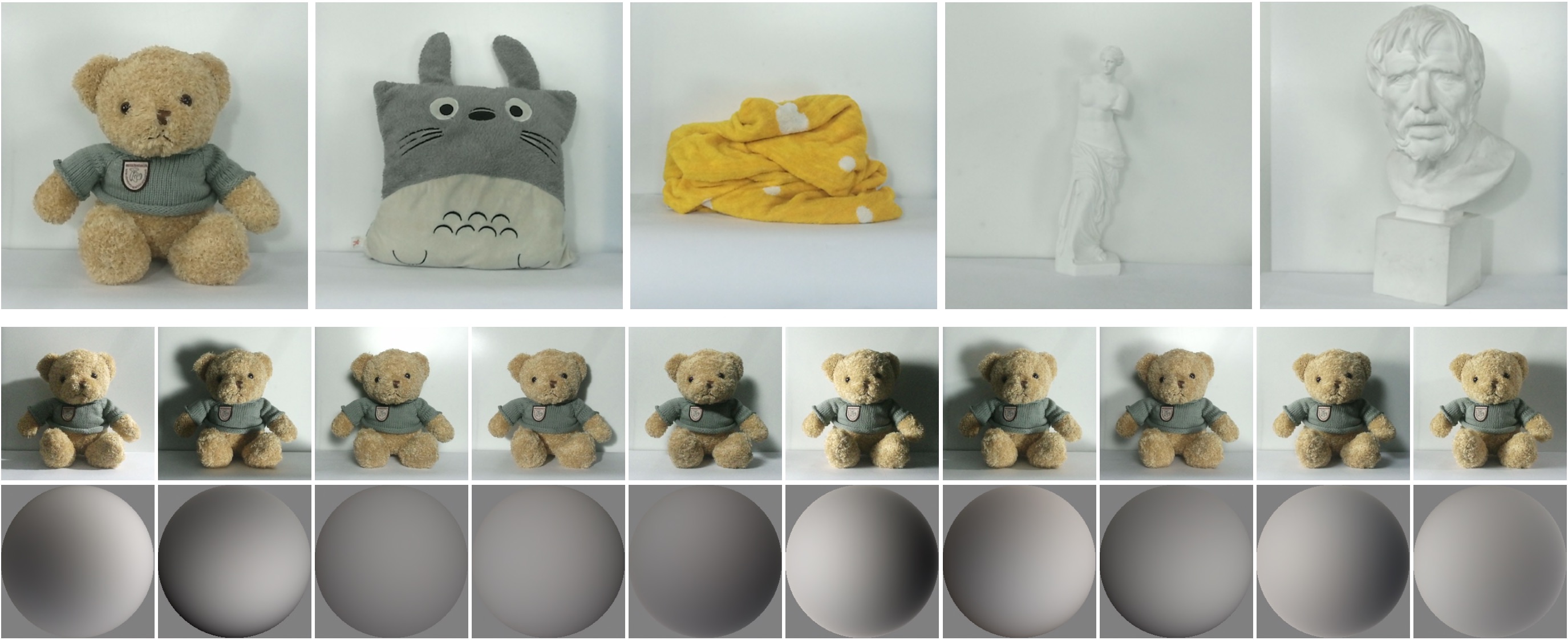}
  \caption{A gallery of the dataset for the shape study experiment. Top row: 5 of the 10 objects of the dataset. Middle row: the 10 images under different lighting conditions of the teddy bear. Bottom: visualization of the estimated ground truth SH light of the corresponding images in the second row.}\label{fig:dataset}
\end{figure*}

We also need a new dataset for this evaluation because existing intrinsic image datasets (e.g. the MIT intrinsic image
dataset) do not have Kinect depths,  and the NYU RGBD dataset has very low resolution that is undesirable for our
relighting purpose. Our new dataset contains 10 example objects. Each example has one diffuse image under a normal
lighting (with no strong directional light) to allow estimation of the model, one aligned depth image captured by a
Kinect-II, and 10 additional images that are taken under the same camera pose but different lighting conditions, and for
each of the 10 images, a diffuse  light probe is placed nearby and photographed for the recovery of the ground truth
lighting  coefficients~\cite{Ramamoorthi:2001:SH}. Figure~\ref{fig:dataset} shows a few examples of the new dataset, and
one example's 10 lighting  images and their estimated lighting.

\begin{figure}[t]
  \includegraphics[width=0.495\linewidth]{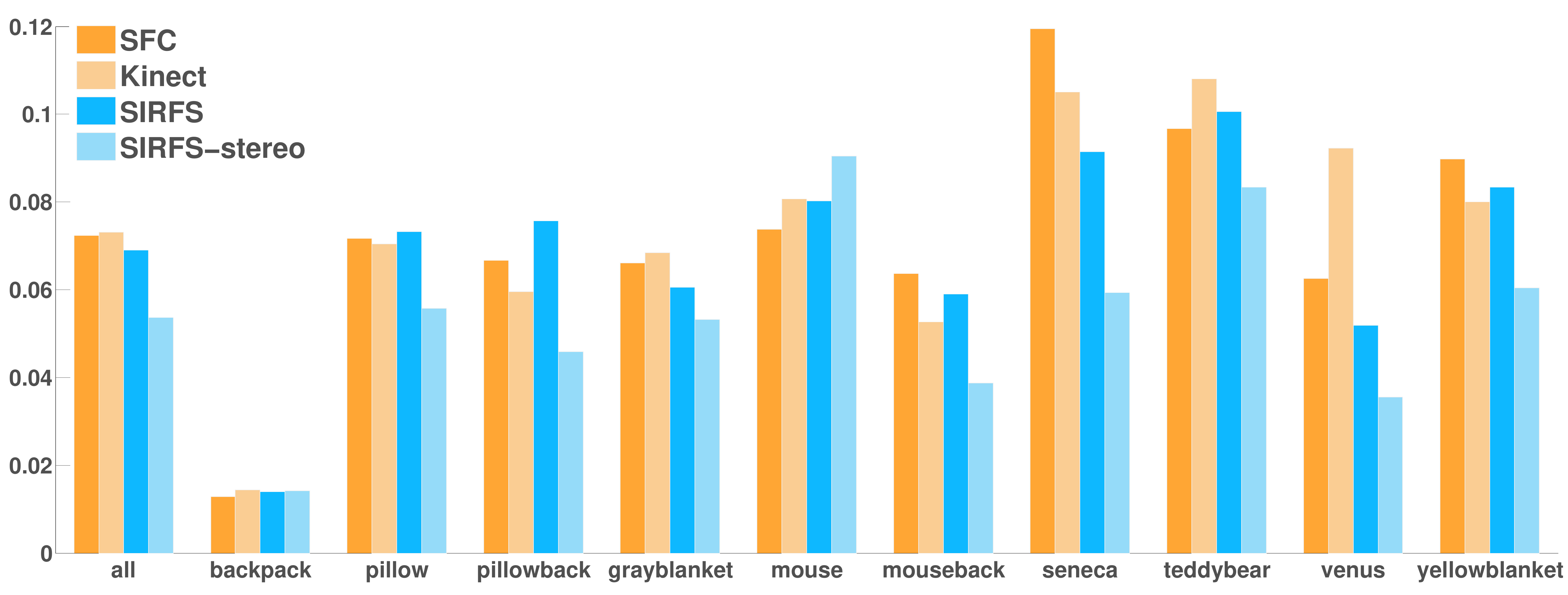}\hfill
  \includegraphics[width=0.495\linewidth]{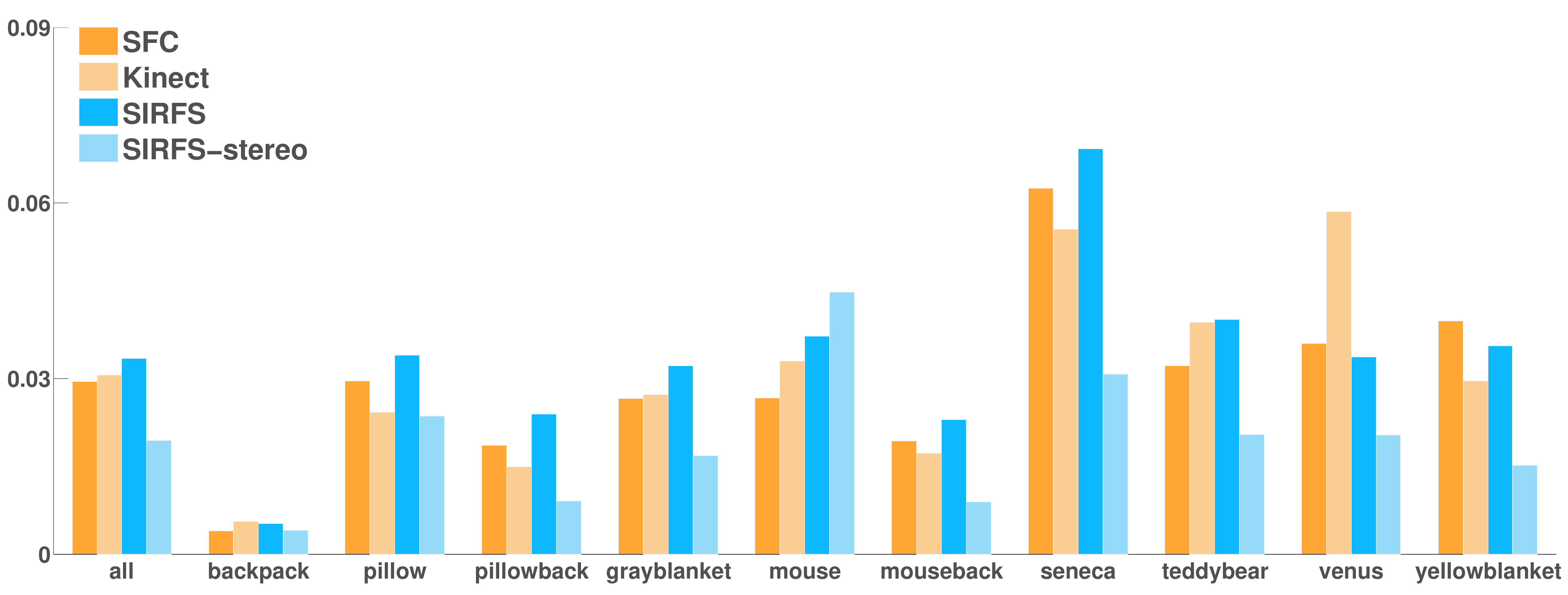}
  \caption{Visualization of the averaged re-rendering error and that of the per-example errors. For each group, the error of each shape configuration is visualized. Top: error of the default setting; Bottom: error of the oracle setting.}\label{fig:shaperes}\vspace{-3mm}
\end{figure}

\vspace{3mm}
\noindent{\bf Experiment}
In the experiment, we are going to build for each object four models using (a) the SFC base shape, (b) the Kinect base shape, (c) the SIRFS base shape and (d) the SIRFS-stereo base shape. For each option, the rest of the model (albedo and the two detail layers) is derived as described in section~\ref{sec:createmodel}. We then relight each object with its four models under the 10 lighting conditions. So for each object and each lighting, we have one ground truth image from the dataset, and four relighting results that are different only because of the based shape. Comparing the relighting results to the ground truth image gives us the re-rendering MSE error (eq.~\ref{eq:MSEmetric}) for each model; the averaged errors are displayed in Table~\ref{tab:shaperes}. In the table, the oracle setting is the same as that described in section~\ref{subsec:rerender_error}. The default setting is different from the regression method in section~\ref{subsec:rerender_error}. Instead, we use the default weight 1 for $w_p$ and $w_g$, and then compute the MSE up to a free scaling factor.
Figure~\ref{fig:shaperes} visualizes the overall and per-instance errors of Table~\ref{tab:shaperes}.

\begin{table}[t]
  \centering
  \begin{tabular}{c||cccc}
  Method & SFC (ours) & Kinect & SIRFS & SIRFS-stereo \\
  \hline\hline
  default & 0.0723 & 0.0732 & 0.0690 & 0.0537\\
  oracle & 0.0295 & 0.0304 & 0.0334 & 0.0194
  \end{tabular}\vspace{2mm}
  \caption{Relighting MSE error on 4 different base shapes: shape from contour, Kinect depth, shape from SIRFS and  shape from SIRFS stereo.
  }\label{tab:shaperes}
\end{table}

\noindent{\bf Discussion}
 Our SFC base shape is comparable to the Kinect and SIRFS shape methods in terms of re-rendering error in the default setting, and is slightly better in the oracle setting. There is an important result here.  Our simple base shape estimate is not the major performance barrier in this task --
more complex estimates of the base shape do not outperform it.  There are two possible explanations: first, the
alternative shapes (Kinect and SIRFS) are not that good either, so the simple and stable SFC shape representation has
certain advantage; second, the detail layers are moderately successful at compensating for missing details, so the
smooth SFC shape only need to take care  of the smooth shading. On the other hand, the SIRFS-stereo shape gives
significantly higher performance, which  means the current model (with SFC base shape) still has considerable space to improve.

The Kinect shape does not work particularly well. This is because we used the Kinect-II, which is a time-of-flight depth sensor rather than the structured light sensor (Kinect-I). The time-of-flight sensor works best for foreground/background separation (i.e., for hand/body tracking), but it has very low depth disparity on objects. Kinect-I might give better depth disparity but we did not use it because it is out dated and it has very low RGB resolution. A more accurate object-level depth sensor is certainly going to boost the relighting performance of our model from the current state, just like the stereo shape did. And note we only need the depth sensor to acquire depth in the coarse scale, because we still have the detail layers for the rest.

In Figure~\ref{fig:shaperes}, it shows the \emph{backpack} has much lower errors than the average errors. This is because it is a \emph{black} backpack, so the overall image values are small, which proportionally affects the errors, because the re-rendering MSE is dependent on the target image (if we scale a target image by a factor of two, the errors is to increase by about the same factor). Therefore, it does not make much sense to compare errors across objects or datasets. Instead, it is the relative errors that matter, as is shown in Figure~\ref{fig:shaperes}.

It is worth noting that MSE is not geared toward \emph{visual realism} of relighting results (non-linearity of visual perception; image features take little weights, etc.). So, we further conduct a set of user studies to measure the subjective user ratings to our relighting results.

\vspace{0mm}
\subsection{User study}\label{subsec:userstudy}
\begin{figure}[t]
\vspace{3mm}
\centering{
\begin{overpic}[width=.32\linewidth,clip,trim=36 50 100 40]{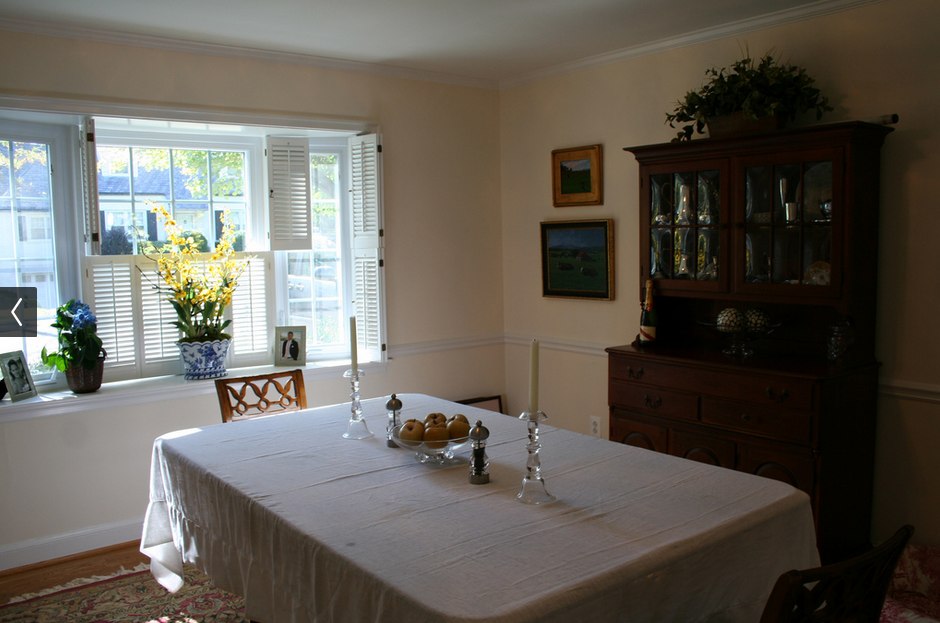}\put(93,2){\colorbox{white}{\small A1}}
\end{overpic}
\begin{overpic}[width=.32\linewidth]{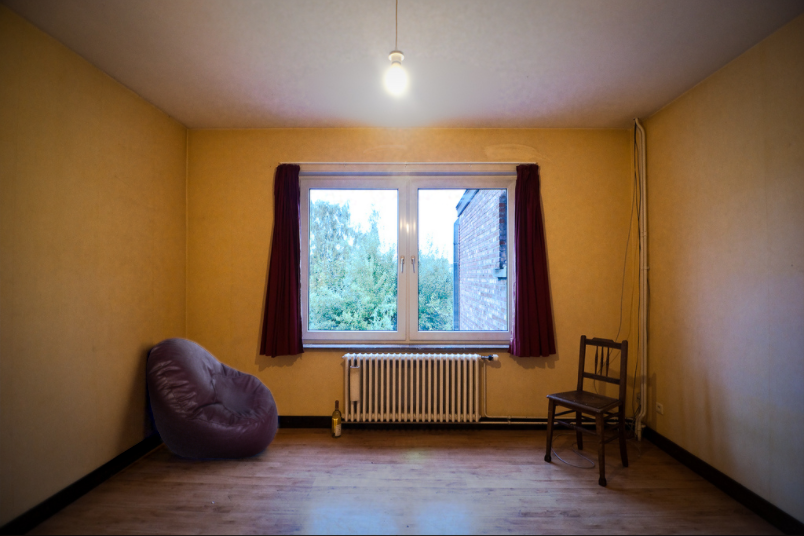}\put(93,2){\colorbox{white}{\small B1}}
\end{overpic}
\begin{overpic}[width=.32\linewidth,clip,trim=0 0 0 33]{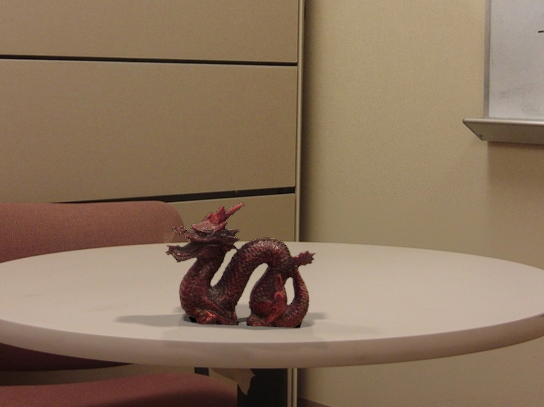}\put(93,2){\colorbox{white}{\small C1}}
\end{overpic}
}

\vspace{.5mm}
\centering{
\begin{overpic}[width=.32\linewidth,clip,trim=0 0 0 68]{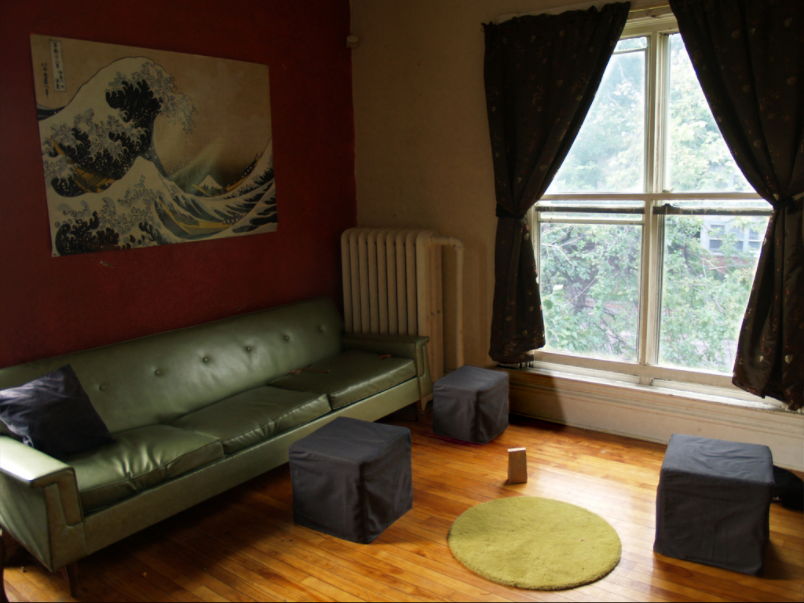}\put(93,2){\colorbox{white}{\small A2}}
\end{overpic}
\begin{overpic}[width=.32\linewidth]{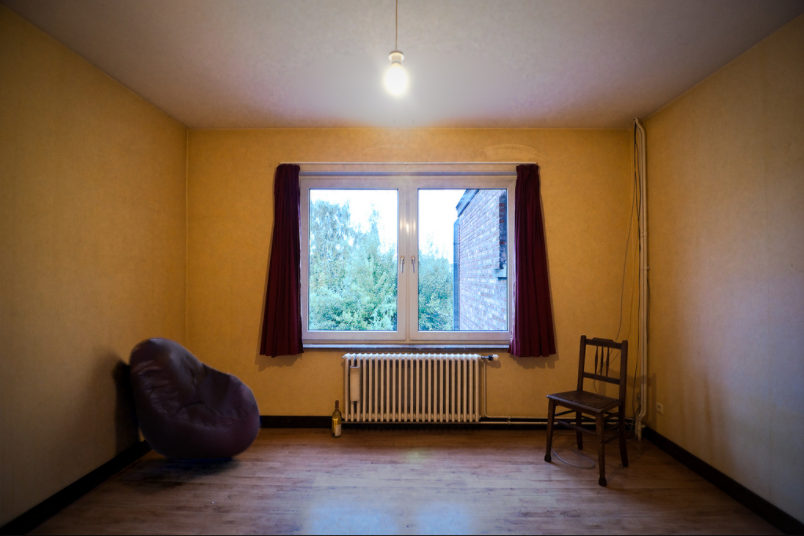}\put(93,2){\colorbox{white}{\small B2}}
\end{overpic}
\begin{overpic}[width=.32\linewidth,clip,trim=0 0 0 33]{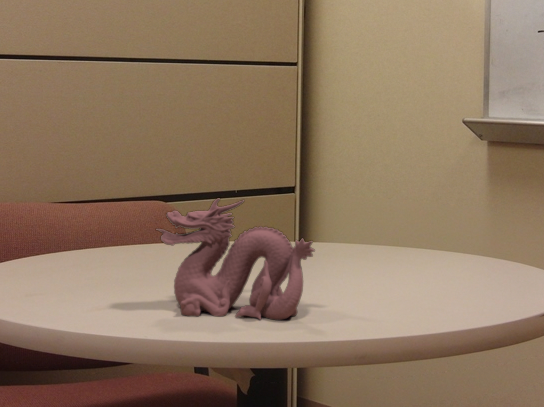}\put(93,2){\colorbox{white}{\small C2}}
\end{overpic}
}\vspace{1mm}
\caption{Example trial pairs from our user study. The left column shows an insertion result by our method and a real image (task 1). The middle column shows that from our method and the method of Barron and Malik (task 3). And the right column shows that from our method and the method of Karsch et al.~\protect{\cite{Karsch:2011}}. Users were instructed to choose the picture from the pair that looked the most realistic. For each row, which image would you choose? Best viewed in color at high-resolution.
\protect\begin{turn}{180} \leftline{\small \it Our results: A2, B1, C1 (ottoman inserted near window in A2);}\protect\end{turn}}
\label{fig:studypics}
\end{figure}

In the study, each subject is shown series of two-alternative forced choice tests and chooses between each pair which he/she feels the most realistic. We tested five different tasks: \RB{(1) our method against real images, (2) our method against Barron and Malik~\cite{Barron:2012B}, and (3) our method against Karsch et al.~\cite{Karsch:2011}, and (4) our method with controlled number of detail layers. The last task shows both detail layers help make more visually realistic result. The other three tasks show the advantage of our result over that of Barron and Malik~\cite{Barron:2012B} and Karsch et al.~\cite{Karsch:2011}. Figure~\ref{fig:studypics} shows example image pairs used in the user study.}

%------------------------ User study example images
%------------------------ User study example images

\vspace{2mm}
\noindent{\bf Experiment} For each task, we created 10 different insertion results using a particular method (ours, Barron and Malik, or Karsch et al.
For the results of Barron and Malik, we ensured the same object was inserted at roughly the same location as our results. This was not the case for the results of Karsch et al., as synthetic models were not all available for the objects we chose.
We also collected 10 real scenes (similar to the ones with insertion) for the tasks involving real images. Each subject viewed all 10 pairs of images for one but only one of the five tasks.
For the 10 results by our method, the detail layer weights were manually selected (it is hard to apply the regression model as in Section~\ref{subsec:rerender_error} to the real scene illuminations) while the other two methods do not have such options.

We polled 100-200 subjects using Mechanical Turk for each task. In an attempt to avoid inattentive subjects, each task also included four ``qualification'' image pairs (a cartoon picture next to a real image). Subjects who incorrectly chose any of the four cartoon picture as realistic were removed from our findings (6 in total, leaving 294 studies with usable data). At the end of the study, we showed subjects two additional image pairs: a pair containing rendered spheres (one a physically plausible, the other not), and a pair containing line drawings of a scene (one with proper vanishing point perspective, the other not). For each pair, subjects chose the image they felt looked most realistic. Then, each subject completed a brief questionnaire, listing demographics, expertise, and voluntary comments.

These answers allowed us to separate subjects into subpopulations: {\bf male/female}, {\bf age $<25$ / $\geq 25$}, whether or not the subject correctly identified both the physically accurate sphere {\it and} the proper-perspective line drawing at the end of the study ({\bf passed/failed perspective-shading (p-s) tests}), and also {\bf expert/non-expert} (subjects were classified as experts only if they passed the perspective-shading tests {\it and} indicated that they had expertise in art/graphics). We also attempted to quantify any learning effects by grouping responses into the {\bf first half} (first five images shown to a subject) and the {\bf second half} (last five images shown).

\noindent{\bf Results and discussion}
\RB{
In the first task, the user study results show that subjects confused our insertion result with a real image in 44\% of 1040 viewed pairs (task 1, see table~\ref{tab:study_vs_real}); an optimal result would be 50\%. We also achieve better confusion rates than the insertion results of Barron and Malik~\cite{Barron:2012B} (42\%), and perform well ahead of the method of Barron and Malik in a head-to-head comparison (task 2, Fig.~\ref{fig:study_vs_barron_karsch} left), as well as a head-to-head comparison with the method of Karsch et al.~\cite{Karsch:2011} (task 3, Fig.~\ref{fig:study_vs_barron_karsch} right). In the last task, users consistently preferred insertion results with more detail layers applied (Figure~\ref{fig:controlled_details}).

Table~\ref{tab:study_vs_real} demonstrates how well images containing inserted objects (using either our method or Barron and Malik) hold up to real images (tasks 1).  We observe better confusion rates (e.g. our method is confused with real images more than the method of Barron and Malik) overall and in each subpopulation except for the population who failed the perspective and shading tests in the questionnaire.

%------------------------ User study results: ours-real vs Barron-real
\begin{table}[!t]
\vspace{1mm}
\centerline{
\begin{tabular}{ l || c c c }
%\hline
Subpopulation & \#trials & ours (\%) & B \& M~\cite{Barron:2012B} (\%) \\ \hline\hline
all & 1040 & \bf 44.0$\pm$1.5 & 41.8$\pm$1.6\\ \hline
expert & 200 & \bf 43.5$\pm$3.4 &  36.2$\pm$3.7\\
non-expert & 840 & \bf 44.2$\pm$1.7 & 42.9$\pm$1.8\\ \hline
passed p-s test & 740 & \bf 44.2$\pm$1.8 &  40.5$\pm$2.1\\
failed p-s test & 300 &  43.7$\pm$2.7 & \bf 43.9$\pm$2.5\\ \hline
male & 680 & \bf 44.7$\pm$1.9 &  42.6$\pm$1.8\\
female & 360 & \bf 42.8$\pm$2.4 &  39.0$\pm$3.5\\ \hline
age $\leq$ 25 & 380 & \bf 43.2$\pm$2.5 & 41.7$\pm$2.5\\
age $>$25 & 660 & \bf 44.5$\pm$1.9 &  41.9$\pm$2.0\\ \hline
1st half & 520 & \bf 45.6$\pm$2.2 &  43.0$\pm$2.3\\
2nd half & 520 & \bf 42.5$\pm$2.0 &  41.8$\pm$2.1\\
%\hline
\end{tabular}}
\vspace{2mm}
\caption{Fraction of times subjects chose an insertion result over a real image in the study. Overall, users confused our insertion results with real pictures 44\% of the time, while confusing the results of Barron and Malik with real images 42\% of the time. Interestingly, for the subpopulation of ``expert'' subjects, this difference became more pronounced (44\% vs 36\%). Each cell shows the mean standard deviation.
}\label{tab:study_vs_real}
%\vspace{-5mm}
\end{table}
%------------------------ User study results: Ours-real vs Barron-real

We also compared our method and the method of Barron and Malik head-to-head by asking subjects to choose a more realistic image when shown two similar results side-by-side (Fig.~\ref{fig:studypics} middle column). Figure~\ref{fig:study_vs_barron_karsch} summarizes our findings. Overall, users chose our method as more realistic in a side-by-side comparison on average 60\% of the time in 1000 trials. In all subject subpopulations, our method was preferred by a large margin to the method of Barron and Malik; each subpopulation was at least two standard deviations away from being ``at chance'' (50\% -- see the red bars and black dotted line in Fig.~\ref{fig:study_vs_barron_karsch}). Most interestingly, the expert subpopulation preferred our method by an even greater margin (66\%), indicating that our method may appear more realistic to those who are good judges of realism.

%------------------------ User study results: ours vs Barron /  Karsch
\begin{figure}[t]
\begin{minipage}{0.495\linewidth}
\centering{\tiny Fraction of times users chose our insertion result over that of Barron and Malik} \\
\includegraphics[trim=0 0 0 16,clip,width=\linewidth]{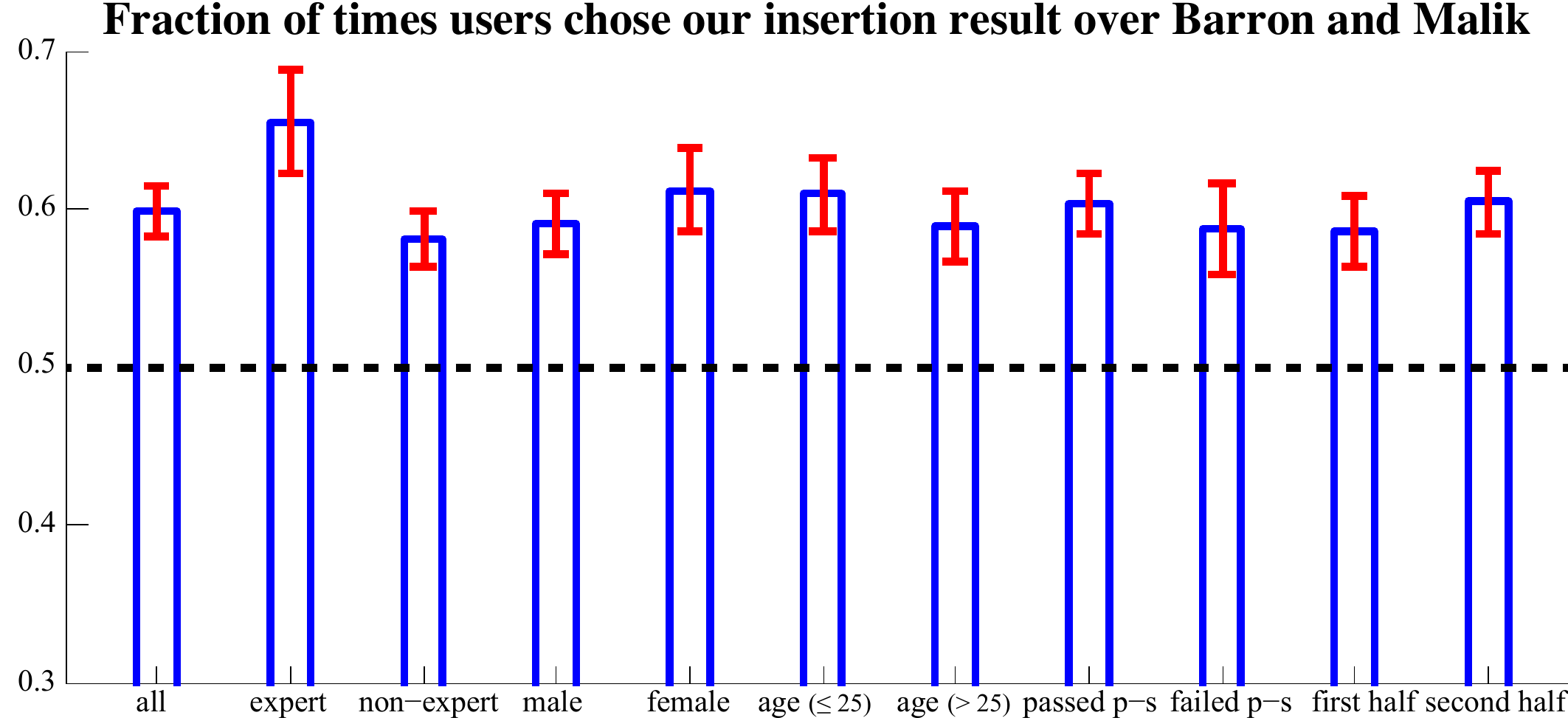}
\end{minipage}
\begin{minipage}{0.495\linewidth}
\centering{\tiny Fraction of times users chose our insertion result over that of Karsch et al.} \\
\includegraphics[trim=0 0 0 15,clip,width=\linewidth]{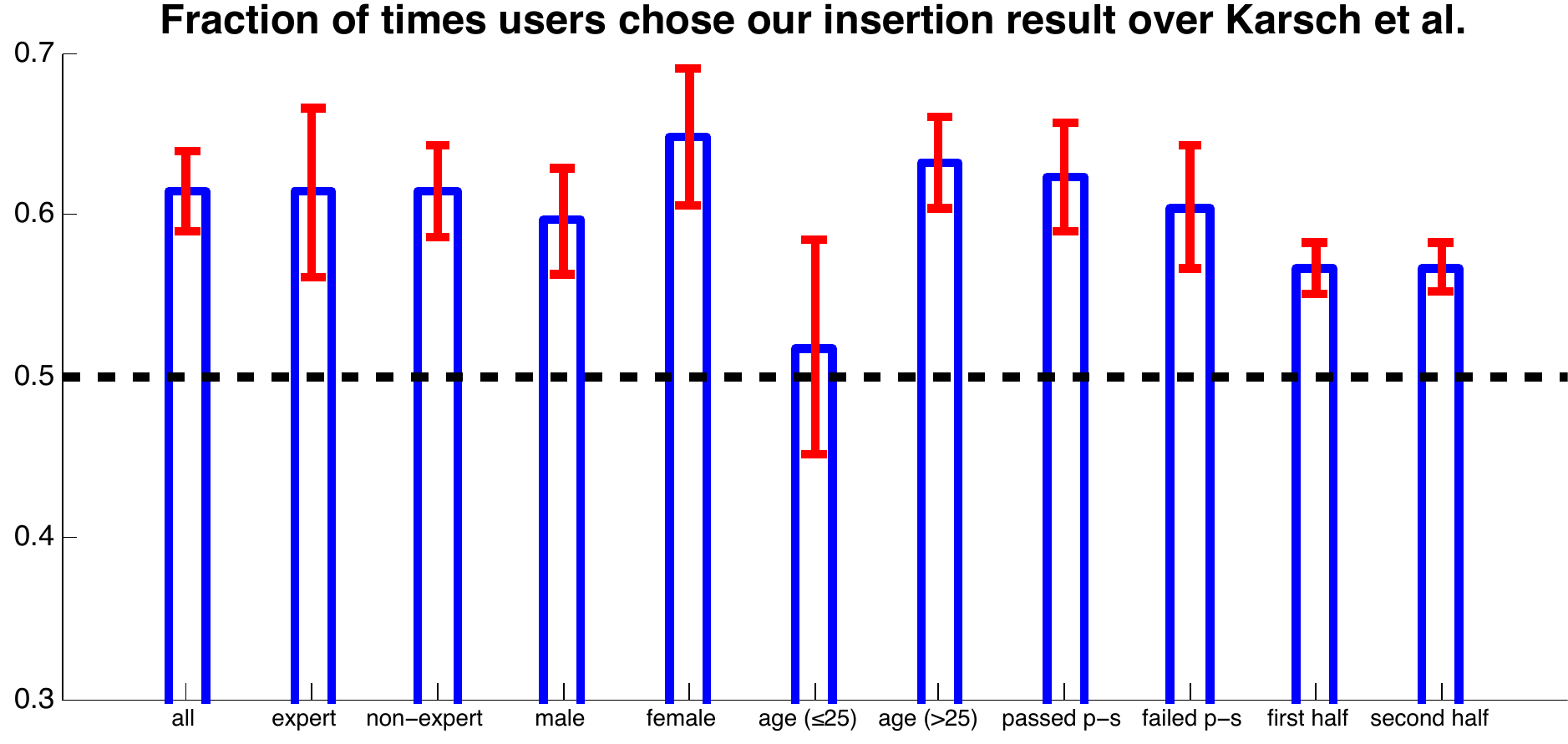}
\end{minipage}
%\centering{
%\includegraphics[width=0.49\linewidth]{images/userstudy/us_vs_barron.pdf}
%\includegraphics[width=0.49\linewidth]{images/userstudy/us_vs_karsch.pdf}
%}
\caption{Left: Comparison of our results against that by the method of Barron and Malik. our results were chosen as more realistic in 60\% of the trials ($N=1000$). For all subpopulations, our results were preferred well ahead of the other as well. All differences to the dotted line (equal preference) are greater than two standard deviation. The ``expert'' subpopulation chose our insertion results most consistently. Right: Comparison against the method of Karsch et al. The advantage our method holds over that of Karsch et al. is similar to the advantage over Barron and Malik. Number of trials = 1840.
}\label{fig:study_vs_barron_karsch}
\end{figure}

%------------------------ User study results: our self-comparison
\begin{figure}[t]
\begin{minipage}{0.495\linewidth}
\centering{\tiny Fraction of times users chose result with one detail layer vs. that with no detail layer} \\
\includegraphics[trim=0 0 0 25,clip,width=\linewidth]{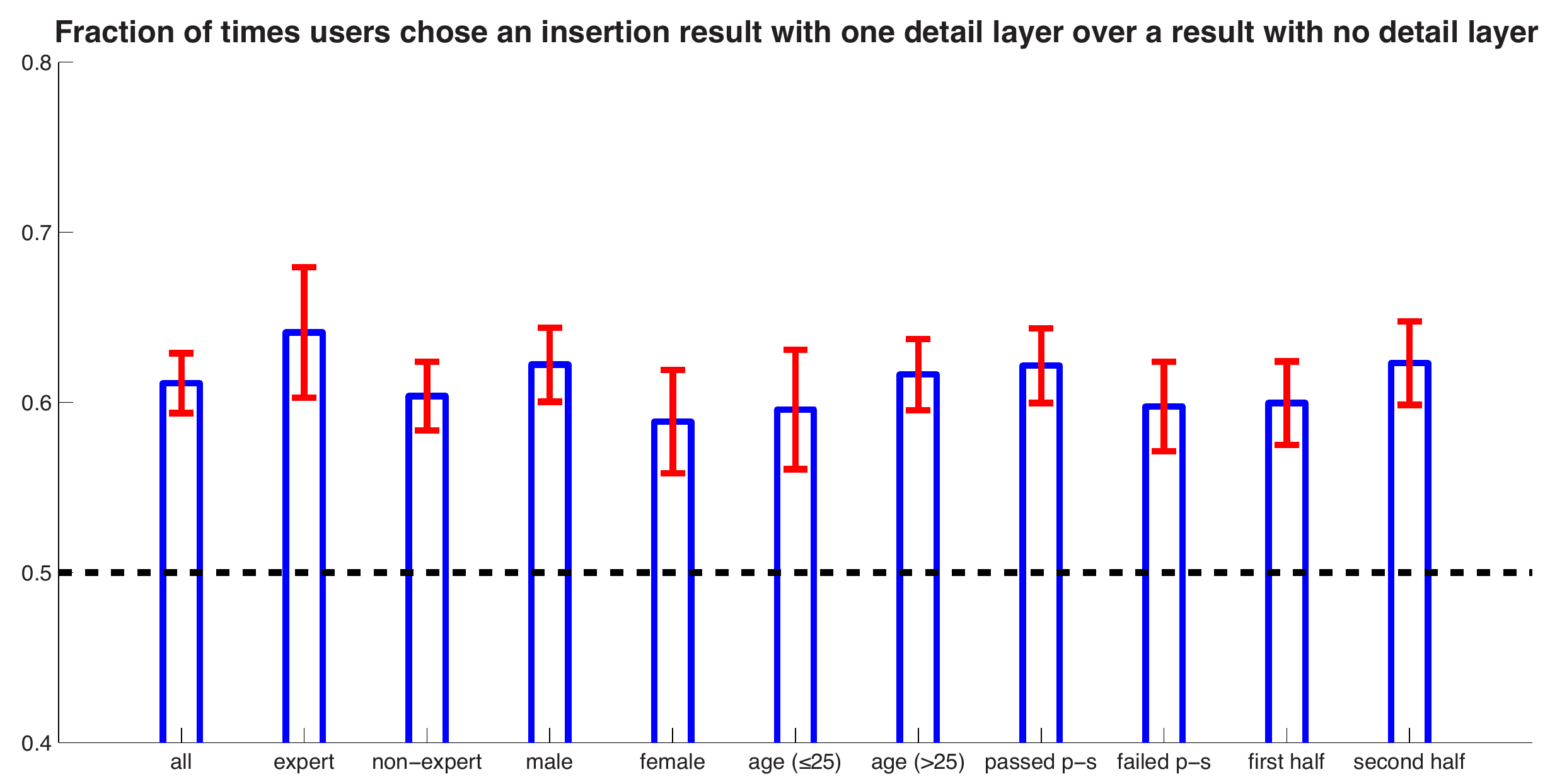}
\end{minipage}
\begin{minipage}{0.495\linewidth}
\centering{\tiny Fraction of times users chose result with two detail layers vs. that with one layer} \\
\includegraphics[trim=0 0 0 20,clip,width=\linewidth]{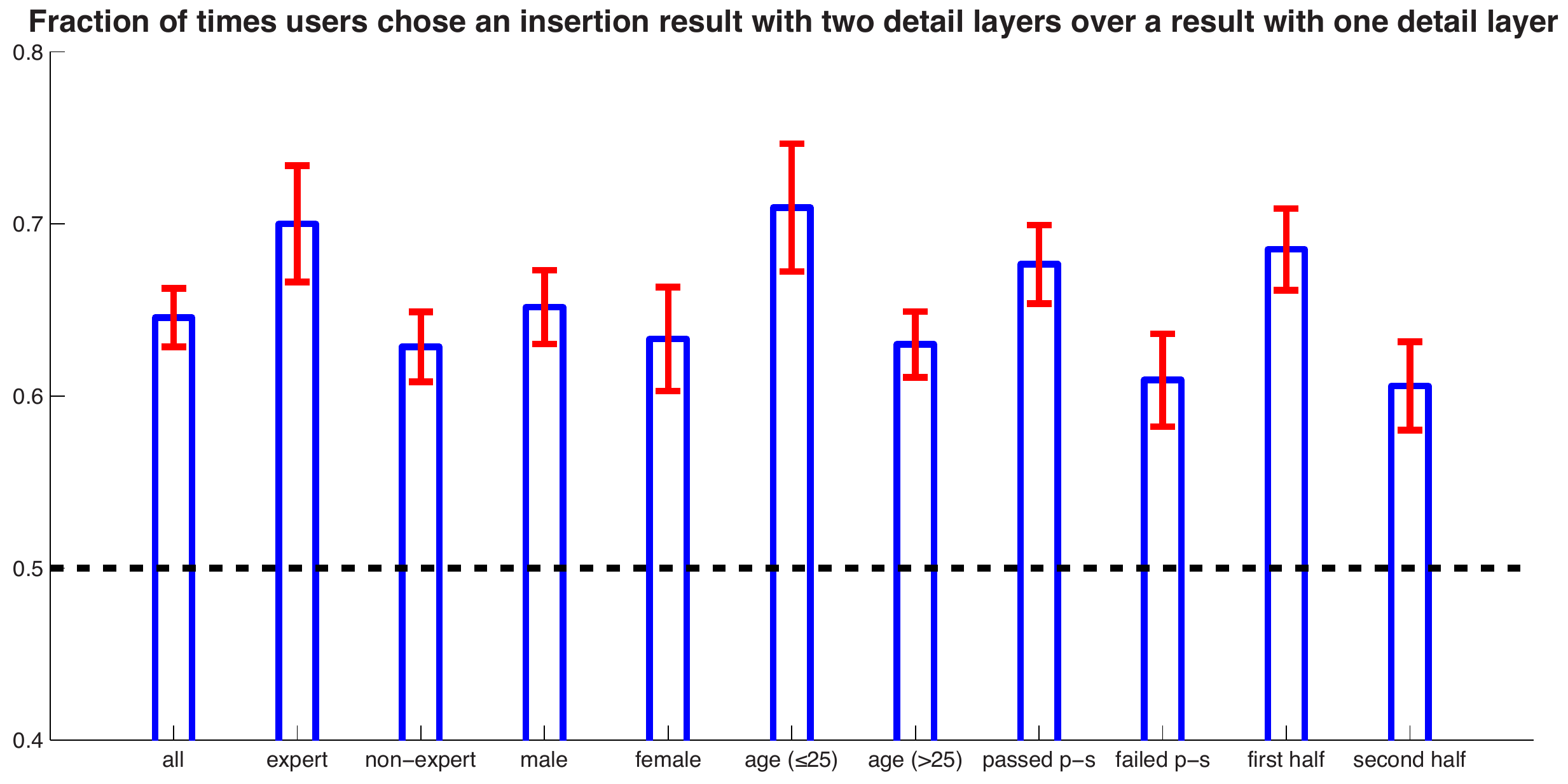}
\end{minipage}
\caption{User rating on our results with controlled number of detail layers. Users consistently prefer insertion results with more detail layers. In the first figure, users prefer results with the parametric detail over that with no detail layer applied in 61\% of 764 viewed pairs. In the second one, users prefer results with both parametric detail and geometric detail over that with only the parametric detail in 65\% of 756 viewed pairs.
}\label{fig:controlled_details}
\end{figure}

Karsch et al.~\cite{Karsch:2011} performed a similar study to evaluate their 3D synthetic object insertion technique, in which subjects were shown similar pairs of images, except the inserted objects were synthetic models. In their study, subjects chose the insertion results only 34\% of the time, much lower than the two insertion methods in this study, a full 10 points lower than our method and 8 points lower than the method of Barron and Malik. While the two studies were not conducted in identical setting, the results are nonetheless intriguing. We postulate this large difference is due to the nature of the objects being inserted: we use {\it real} image fragments that were formed under real geometry, complex material and lighting, sensor noise, and so on; they use 3D models for which photorealism can be extremely difficult to model. By inserting image fragments instead of 3D models, we gain photorealism in a data-driven manner (Fig.~\ref{fig:studypics} C1 versus C2). This postulation is validated by our comparison in task 5. For all but one subpopulations, our results were preferred by a large margin (Fig.~\ref{fig:study_vs_barron_karsch} right).

Figure~\ref{fig:controlled_details} displays the result of the last task. It demonstrates the model makes better insertion results as more detail layers are applied. Overall, users preferred insertion results with detail 1 over that without detail composition in 61\% of 764 viewed image pairs, and preferred results with both detail layers over that with only detail 1 in 65\% of 756 viewed pairs. Consistent results were shown in all subpopulations.
}

\begin{figure}[t]
\centering{
\subfigure[Input]{
\includegraphics[height=.9in]{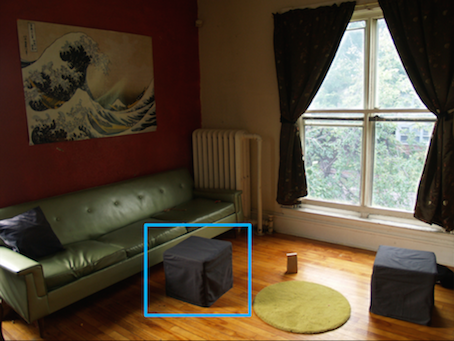}\hspace{0mm}}
\subfigure[Shape]{
\includegraphics[height=.9in]{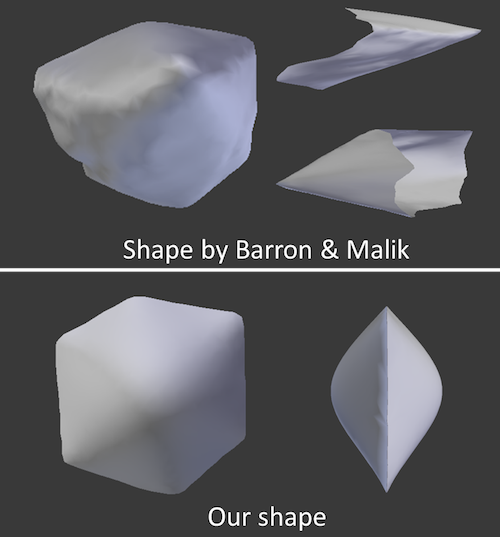}\hspace{0mm}}
\subfigure[B \& M]{
\includegraphics[height=.9in]{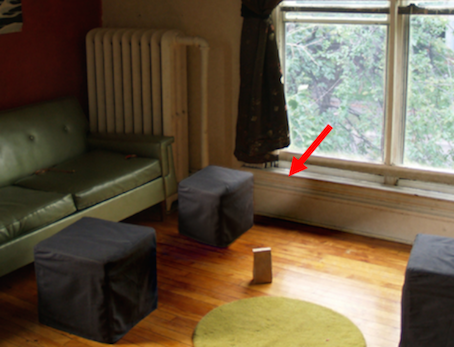}\hspace{0mm}}
\subfigure[Our result]{
\includegraphics[height=.9in]{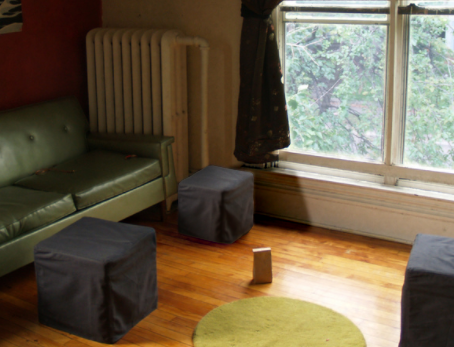}}
}
\caption{In this example, we built models from the cube in the input image (cyan box) and inserted it back into the scene. Sophisticated SfS methods (in this case, Barron and Malik~\protect\cite{Barron:2012B}) can have large error and unstable boundaries that violates the generic view assumption. For object insertion, lighting {\it on} the object is important, but it is equally important that cast shadows and interreflected light look correct; shape errors made by complex SfS methods typically exacerbate errors both {\it on} and {\it around} the object (see cast shadows in c). Our shape is simple but behaves well in many situations and is typically robust to such errors (d). Best viewed in color at high-resolution.}
\label{fig:cube}
\vspace{0mm}
\end{figure}

\section{Conclusion and future work}
\Ra{

We have presented an approximate shading model and the accompanying algorithms for building the corresponding object model from single image, and a relighting system that supports image-to-image object insertion with an interactive interface for user control. The shading model is based on psychological findings of human visual perception, and therefore distinguishes from existing physically-based shading approach. In the end, the model takes a hybrid of physically-based graphics rendering and image-based detail composition. The detail components enable the object model to accommodate surface mesostructure for which 3D mesh representation is limited, and allows visual effect of such fine-scale structure to be modeled and transferred directly from image to image with illumination change.

The system can be improved in several directions. First, the object is constrained to be relit from roughly the same camera direction. A slight view angle perturbation is manageable, but larger view angle change is to expose the back of the object, because the object model is constructed from a single view and not as a 3D model. Integrating geometric stereo techniques to recover full 3D object model would be an interesting next step. Second, the core part of the model -- the two detail layers -- can be further improved. One important direction is to extend them to a larger number of components for finer modeling granularity and control. It remains an open question as to explore more principled ways of defining such detail series other than the current handcrafted approach. Another possible direction of future work is to focus to the very important case of human face. We can make use of category specific prior for better intrinsic decomposition, exploit parametric face model for better depth recovery, and, possibly, learn more powerful shape or material extraction models from data -- if a suitable face dataset is publicly available.}

\RB{
Another direction is the integration of the widely adopted deep neural network to our system. We have seen exciting improvements in intrinsic decomposition~\cite{narihira2015direct,DBLP:KimPSL16,lettry2016darn}, where both end-to-end trained convolutional neural network and adversarial training with a CNN-based generator/discriminator network are proved effective. We have also seen the reconstruction of face normal or depth, or joint reconstruction of face normal, albedo and illumination using the convolutional encoder-decoder framework~\cite{Trigeorgis_2017_CVPR,Richardson_2017_CVPR,Shu_2017_CVPR}. In addition, Soltani et al.~\cite{Soltani_2017_CVPR} propose to synthesize 3D shapes from depth maps and silhouettes using a deep \emph{generative} network. Deshpande et al.~\cite{Deshpande_2017_CVPR} propose to generate diverse color fields from grayscale images using variational autoencoders. The key of the above mentioned problems can be viewed as an analysis (i.e. representation learning) process followed with a synthesis (i.e. generation) process; and deep neural networks are the most convenient computational infrastructure and toolset for this problem up-to-date.  Yet the key component of our system -- the relighting process -- is not seen in neural network-based solutions so far. It remains as an open question whether deep neural networks will be a viable approach. 
The challenges may include in the requirement of high resolution visual outputs, the trade-offs between diversity and physical correctness, and the network's adaptability to unseen geometry or illumination patterns.

%Soltani et al. 
%CNN for intrinsic decomposition
%VAE for colorization (Learning Diverse Image Colorization; cvpr17)
%Face disentanglement (Neural Face Editing With Intrinsic Image Disentangling; cvpr17)
%CVAE-GAN: Fine-Grained Image Generation through Asymmetric Training (ICCV17)

%End-to-End train?
%Conditional generative model.

%Possibility:

%Challenges: 
%diversity vs. physical correctness (controversial)
%high resolution
%traditional solutions readily available.
}

\vspace{2mm}
{\noindent\bf Acknowledgements }
DAF is supported in part by Division of Information and Intelligent Systems (US)
(IIS 09-16014), Division of Information and Intelligent Systems
(IIS-1421521) and Office of Naval Research (N00014-10-10934).
ZL is supported in part by NSFC grant No. 61602406, ZJNSF Grant No. Q15F020006 and a special fund from Alibaba -- Zhejiang University Joint Institute of Frontier Technologies.

%\clearpage
{\small
\bibliographystyle{ieee}
\bibliography{zcl}
}

%\input{supp}

%\clearpage \input{supp}

\end{document}